\documentclass[10pt,journal,compsoc]{IEEEtran}

\ifCLASSOPTIONcompsoc
  \usepackage[nocompress]{cite}
\else
  \usepackage{cite}
\fi

\ifCLASSINFOpdf
\else
\fi

\usepackage{url}            % simple URL typesetting
\usepackage{booktabs}       % professional-quality tables
\usepackage{amsfonts}       % blackboard math symbols
\usepackage{nicefrac}       % compact symbols for 1/2, etc.
\usepackage{microtype}      % microtypography
\usepackage{xcolor}         % colors
\usepackage{graphicx}
\usepackage{amsmath,amssymb}
\usepackage{verbatim}
\usepackage{multirow}
\usepackage{marvosym}
\usepackage{makecell}
\usepackage{bm}
\usepackage{subfigure}
\usepackage{wrapfig}
\usepackage{caption}
\usepackage[marginal]{footmisc}

\newcommand{\mypar}[1]{{\bf #1.}}

\newcommand{\R}{\ensuremath{\mathbb{R}}}

\def\x{\mathbf{x}}

\def\X{\mathbf{X}}

\begin{document}
\title{Dynamic-Group-Aware Networks for Multi-Agent Trajectory Prediction with Relational Reasoning}

\author{Chenxin Xu, Yuxi Wei, Bohan Tang, Sheng Yin, Ya Zhang, Siheng Chen
\IEEEcompsocitemizethanks{\IEEEcompsocthanksitem C. Xu, Y. Wei, S. Yin, Y. Zhang and S. Chen are with the Cooperative Medianet Innovation Center (CMIC) at
Shanghai Jiao Tong University, Shanghai, China. 
E-mail: xcxwakaka, wyx3590236732, Yin.sheng011224, ya\_zhang, sihengc@sjtu.edu.cn 

\IEEEcompsocthanksitem B. Tang is with the Oxford-Man Institute and the Department of Engineering Science, University of Oxford, Oxford OX2 6ED, UK. E-mail: bohan.tang@eng.ox.ac.uk. 

\IEEEcompsocthanksitem Corresponding authors are Siheng Chen and Ya Zhang. \protect \\
}
}

% \markboth{Journal of \LaTeX\ Class Files,~Vol.~14, No.~8, August~2015}%
% {Shell \MakeLowercase{\textit{et al.}}: Bare Demo of IEEEtran.cls for Computer Society Journals}

\IEEEtitleabstractindextext{%
\begin{abstract}
Demystifying the interactions among multiple agents from their past trajectories is fundamental to precise and interpretable trajectory prediction. However, previous works mainly consider static, pair-wise interactions with limited relational reasoning. To promote more comprehensive interaction modeling and relational reasoning, we propose~\texttt{DynGroupNet}, a dynamic-group-aware network, which can i) model time-varying interactions in highly dynamic scenes; ii) capture both pair-wise and group-wise interactions; and iii) reason both interaction strength and category without direct supervision.  Based on~\texttt{DynGroupNet}, we further design a prediction system to forecast socially plausible trajectories with dynamic relational reasoning. The proposed prediction system leverages the Gaussian mixture model, multiple sampling and prediction refinement to promote prediction diversity, training stability and trajectory smoothness, respectively. Extensive experiments show that: 1)~\texttt{DynGroupNet} can capture time-varying group behaviors, infer time-varying interaction category and interaction strength during trajectory prediction without any relation supervision on physical simulation datasets; 2)~\texttt{DynGroupNet} outperforms the state-of-the-art trajectory prediction methods by a significant improvement of 22.6\%/28.0\%, 26.9\%/34.9\%, 5.1\%/13.0\% in ADE/FDE on the NBA, NFL Football and SDD datasets and achieve the state-of-the-art performance on the ETH-UCY dataset.

\end{abstract}

}

\maketitle

\IEEEdisplaynontitleabstractindextext

\IEEEpeerreviewmaketitle

% \tableofcontents

\IEEEraisesectionheading{\section{Introduction}\label{sec:introduction}}

\IEEEPARstart{M}ulti-agent trajectory prediction aims to predict future trajectories of multiple agents conditioned on their past movements. This task is critical to numerous real-world applications, such as autonomous driving \cite{gao2020vectornet,liang2020learning,casas2018intentnet}, industrial robotics \cite{demiris2007prediction,forestier2016autonomous}, human behavior understanding \cite{li2019actional,li2021symbiotic} and surveillance \cite{gaur2011string}, serving as a foundation to bridge the knowledge of the past and the action for the future.

In the prediction, at least three factors would affect each agent's dynamics: ego momentum, instantaneous intent, and social influence from the other agents. The first factor has been well studied {\cite{sutskever2014sequence}}; the second factor is unpredictable; and the third factor is an emerging research topic and the focus of this work. To demystify the social influence, we need to model and reason the interactions among agents based on their past spatio-temporal states, potentially leading to precise and interpretable trajectory prediction.

% \begin{figure}[t] 
% \centering
% \includegraphics[width=0.46\textwidth]{latex/img/jewel.pdf}
% \vspace{-3mm}
% \caption{\small GroupNet captures both pair-wise and group-wise interactions among multiple agents from their trajectories  and infers the interaction category and strength for each interaction group.}
% \label{fig:jewel}
% \vspace{-17pt}
% \end{figure}

A lot of efforts have been made to model social interactions. Concretely, traditional approaches use handcrafted rules \cite{helbing1995social,antonini2006discrete} like setting constraints about attractive forces and repulsive forces for modeling leader following and collision avoiding behavior. Recently, many researchers tend towards deep learning tools for a more general interaction modeling mainly through three mechanisms: spatial-centric mechanism, attention mechanism and graph-based mechanism. The spatial-centric mechanism \cite{chou2018predicting,deo2018convolutional,zhao2019multi,bansal2018chauffeurnet,casas2018intentnet} represents agents' trajectories in a unifying spatial domain and use the spatial relationship to model the interaction. The social or attention mechanism aggregates neighboring agents information in the scene through pooling operation \cite{alahi2016social,gupta2018social}, attention operation \cite{xu2018encoding,mangalam2020not,vemula2018social,sadeghian2019sophie,tang2021collaborative} or transformer structures \cite{yu2020spatio,zhou2022ga,giuliari2021transformer,yuan2021agentformer}. The graph-based mechanism \cite{wu2020connecting,kosaraju2019social,huang2019stgat,hu2020collaborative,mohamed2020social,li2021online,gao2020vectornet,gu2022stochastic} is proposed to explicitly model the interaction between agents through non-grid structures. For example, \cite{huang2019stgat} proposes spatio-temporal graph attention network.  \cite{hu2020collaborative} constructs a directed fully connected graph to model interaction between agents. Some previous works, such as NRI \cite{kipf2018neural} and Evolvegraph \cite{li2020evolvegraph}, not only implicitly model the interaction but also take a step forward towards explicit relational reasoning that explicit infer the interaction relationships like which category of the interaction is. 

\begin{figure}[t] 
\centering
\includegraphics[width=0.48\textwidth]{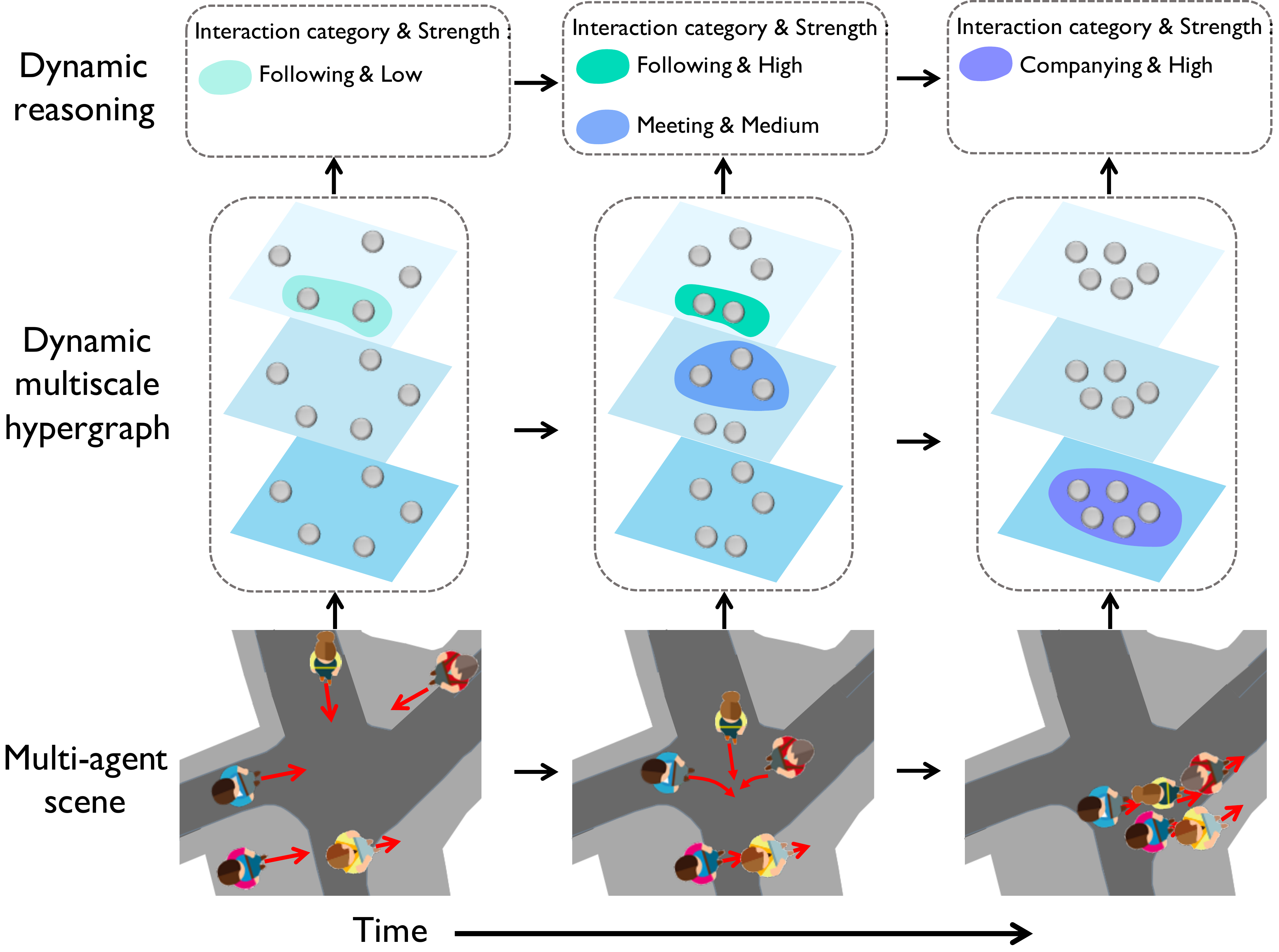}
\vspace{-1mm}
\caption{\small ~\texttt{DynGroupNet} can capture both pair-wise and group-wise time-varying interactions among multiple agents and infer the interaction category and strength for each interaction group across time based on a dynamic multiscale hypergraph.} 
\label{fig:jewel}
\vspace{-2mm}
\end{figure}

However, previous works still have limitations due to the potential complex properties of agent interactions from three aspects. First, in many scenarios, such as on a basketball court, a group of players execute a defensive strategy cooperatively; in the ocean, a school of fish coordinate their movements to evade predators, reflecting collective behaviors. While this kind of group-wise interaction is common, they have rarely been modeled. Second, the interaction relationship would change across time, including both the group belongingness and the interaction inside the group, which is out of consideration by most of methods. Third, from the aspect of reasoning, only interaction categories are explicitly reasoned by current methods, while the intensity level of the agent interaction cannot be reflected, which is common in our world like gravity at various magnitudes.

To further promote more comprehensive interaction modeling and reasoning in trajectory prediction, this work puts efforts on three aspects: interaction capturing, interaction evolving and interaction representation learning. To capture group-wise interactions, we propose a multiscale hypergraph, which consists of a series of hypergraphs to model group-wise interactions with multiple group sizes. Instead of using handcrafted designs, we learn such a multiscale hypergraph topology in a data-driven manner. To capture dynamic interactions, we evolve the multiscale hypergraph to a dynamic multiscale hypergraph on both the topology and representation, modeling the dynamics of group belongingness and interaction inner groups. We propose a recurrent encoding and affinity matrix propagation for the dynamics of multiscale hypergraph topology, and a transformer-based dynamic embedding evolvement integrating previous agents interaction information to current interaction modeling. To reason interactions, we propose a three-element representation format: the neural interaction strength, the neural interaction category and the per-category function, which can reflect the interaction strength and category in an interactive group. Based on neural message passing over the dynamic multiscale hypergraph, we merge this three-element interaction embedding in the representation learning process to obtain a dynamic relational reasoning. Overall, we integrate these three designs and propose~\texttt{DynGroupNet}, a dynamic-group-aware network, to capture time-varying group-wise interactions for better trajectory prediction, see Fig. \ref{fig:jewel} for a sketch. 

To cooperate with the proposed~\texttt{DynGroupNet}, we further propose a general prediction system to predict dynamically-feasible future trajectories. The prediction system takes~\texttt{DynGroupNet} as the core component of the encoding process and predict future trajectories recurrently by setting predicted trajectories to next encoding inputs. To model the uncertainty and diversity
of agent futures, we adopt a Conditional Variational AutoEncoder (CVAE) structure with the bivariate Gaussian Mixture Model (GMM) to model future trajectory's distribution instead of directly regressing results. To reduce the input noise bringing by the inaccurate part of diverse predicted futures and stable the system training, we further propose a multiple sampling training strategy. To enhance the correlation among predicted results at every timestamps which are individually sampled from the GMM, we propose a prediction refinement to achieve a more smooth and feasible future prediction. 

To validate~\texttt{DynGroupNet}, we conduct extensive experiments on both synthetic simulation datasets and real-world datasets. We design the simulation under different scenarios based on physical principles in which the ground truth of the interaction relationship can be accessed. We also validate the effectiveness of the proposed~\texttt{DynGroupNet} along with the prediction system on four real-world datasets: NBA, NFL Football, SDD and ETH-UCY. Experiment results show that 1)~\texttt{DynGroupNet} can capture time-varying group behaviors, infer time-varying interaction strength and interaction category during trajectory prediction without any relation supervision on physical simulation datasets; 2)~\texttt{DynGroupNet} outperforms the state-of-the-art methods by a significant improvement of 22.6\%/28.0\%, 26.9\%/34.9\%, 5.1\%/13.0\% in ADE/FDE on the NBA, NFL Football and SDD datasets and achieve the state-of-the-art performance on the ETH-UCY datasets, reflecting our method's effectiveness; 3) using a dynamic multiscale hypergraph achieves higher performance than using a static multiscale hypergraph, confirming the importance of modeling dynamic interactions; and 4) designs including GMM, multiple sampling and prediction refinement in our prediction system are all beneficial.

The main contributions are as follows:

$\bullet$ We propose~\texttt{DynGroupNet}, a novel dynamic-group-aware network. Based on a dynamic multiscale hypergraph,~\texttt{DynGroupNet} can capture time-varying group-wise interactions at various group sizes and reason the corresponding interaction strength and category.

$\bullet$ We propose a~\texttt{DynGroupNet}-based general trajectory prediction system, which is tightly integrated with Gaussian mixture model, multiple sampling and prediction refinement to achieve diverse prediction, stable training and smooth trajectory.

$\bullet$ We design extensive synthetic simulations based on physical principles.
% in which ground truth of the interaction relationship can be accessed for evaluating relational reasoning. 
Based on the synthetic simulations, we validate the dynamic relational-reasoning ability of~\texttt{DynGroupNet} and show that~\texttt{DynGroupNet} can capture time-varying group behaviors, interaction strength and interaction category in an unsupervised setting. 
    
$\bullet$ We conduct extensive experiments on four real-world datasets to validate that~\texttt{DynGroupNet} achieves state-of-the-art performances on four real-world prediction benchmarks. Our method outperforms previous most state-of-the-art works by 22.6\%/28.0\%, 26.9\%/34.9\%, 5.1\%/13.0\% in ADE/FDE especially on the NBA, Football, SDD datasets.

Compared to our previous work~\cite{xu2022groupnet}, the proposed~\texttt{DynGroupNet} has three major technical improvements:

$\bullet$ From the perspectives of interaction modeling: This work proposes a network based on a dynamic multiscale hypergraph to model and reason the time-varying interactions; while~\cite{xu2022groupnet} only considers a static multiscale hypergraph, which can only model and reason static interactions.  
    
$\bullet$ From the perspectives of temporal dependency: This work proposes a transformer-based dynamic embedding evolvement mechanism to obtain more comprehensive features; while~\cite{xu2022groupnet} does not consider temporal dependency.
    
$\bullet$ From the perspectives of prediction system: This work proposes a bivariate Gaussian mixture model to model the distribution of trajectories, promoting the prediction diversity; while~\cite{xu2022groupnet} directly predicts the coordinates of trajectories. Furthermore, we also design a multiple sampling method and a prediction refinement process to provide the model with more stable training and closer correlation of predicted coordinates in a trajectory.

% Based on those new techniques, we are able to conducts more extensive experiments on both simulation datasets and real-world datasets. The new experiments include:

% $\bullet$ We design more general simulation datasets for dynamic relational reasoning, extending from static relational reasoning.

% $\bullet$ The performance outperforms~\cite{xu2022groupnet} by 21.2\%/33.1\%, 9.6\%/15.7\%, 8.0\%/13.6\% in ADE/FDE on the NBA, SDD and ETH datasets.

% $\bullet$ We add a dataset of NFL Football which contain fierce confrontation and rich interactions. 

% $\bullet$ From the perspectives of experiment: This work conducts more extensive experiments on both simulation datasets and real-world datasets. The performance outperforms~\cite{xu2022groupnet} by \%,\%,\% on NBA, SDD and ETH datasets \Note{CX: because we change the prediction system, should we compare to the cvpr result? or compare to a static multiscale hypergraph result?}.

The rest of the paper is organized as follows: in Section \ref{sec:related work}, we review existing works related to trajectory prediction and relational reasoning. In Section \ref{sec:formulation}, we formulate
the problem of trajectory prediction with dynamic relational reasoning and introduce some mathematical foundations of our model. In Section \ref{sec:dyngroupnet}, we propose a core module~\texttt{DynGroupNet} in our prediction system: dynamic-group-aware network. In Section \ref{sec:system}, we propose a prediction system cooperated with~\texttt{DynGroupNet} and the training objective function. In Section \ref{sec:experiment}, we conduct experiments both in simulated datasets and real-world datasets validating the effectiveness of our method both in relational reasoning and  trajectory prediction. Finally, we draw the conclusion of the paper in Section \ref{sec:conclusion}.

\section{Related Work}
\label{sec:related work}
\textbf{Trajectory prediction.} Traditional approaches use hand-crafted rules and energy potentials \cite{helbing1995social,antonini2006discrete,lee2007trajectory,mehran2009abnormal,morris2009learning,wang2011trajectory,wang2008unsupervised} for dynamic motion prediction. For example, Social Force~\cite{helbing1995social} models pedestrians' behavior by the destination, the repulsive and the attractive effects. In recent years, data-driven approaches are proposed like sequence-to-sequence models \cite{alahi2016social,morton2016analysis,vemula2018social}, which are leveraged to encode trajectories individually for a deterministic prediction. Since the future trajectory is uncertain, researchers begin to propose frameworks to predict multi-model trajectories. For example, generator-discriminator structures are used~\cite{mohamed2020social,huang2019stgat,yu2020spatio,gupta2018social,hu2020collaborative,dendorfer2021mg} with adding noise to generate multiple predictions. \cite{liang2020learning,tang2021collaborative} adapt multi-head output to regress multiple future trajectories. \cite{graber2020dynamic,li2020evolvegraph} utilize a Gaussian mixture distribution to model the future trajectory distribution and estimate the mean and covariance. \cite{bae2022non} proposes a sampling network that directly generates purposive sample sequences. \cite{xu2022remember} proposes a memory mechanism to generate diverse predictions by searching different related memories. Conditional variational autoencoders~\cite{mangalam2020not,lee2017desire,ivanovic2019trajectron,salzmann2020trajectron++,yuan2021agentformer} are frequently used which estimate the parameters of a latent distribution and sample future trajectory features from the distribution. Researchers also put efforts into trajectory prediction in the autonomous driving scenario, which specifically considers the HD map information. For example, \cite{hu2020collaborative,chai2019multipath,liang2020garden,casas2018intentnet} use rasterized encoding methods to rasterize the HD map elements together with agents into an image and use CNNs to encode the image. \cite{gao2020vectornet} propose a vectorized encoding method to directly model structured scene contexts from
their vectorized form. \cite{liang2020learning}
construct a lane graph from raw map data and uses graph convolutions to capture the complex topology of the lane graph. 

No matter how the prediction model and scenario varies, modeling the interactions among multiple agents is fundamental to precise and interpretable trajectory prediction.
Recent works mainly use three mechanisms to model the hidden interactions. The first is spatial-centric mechanism \cite{chou2018predicting,deo2018convolutional,zhao2019multi,bansal2018chauffeurnet,casas2018intentnet}, which represents agent's trajectory in a unifying spatial domain and uses the spatial relationship to model the social interaction. For example, 
\cite{zhao2019multi} utilizes spatial tensor to represent agent interaction. \cite{bansal2018chauffeurnet} encodes the trajectories into rasterized bird-eye view images. The second is social or attention mechanism. \cite{alahi2016social,gupta2018social} use a social mechanism aggregating the neighboring agents’ information to a social representation and broadcast it to each agent thus each agent is aware of the neighboring information. Attention mechanism \cite{xu2018encoding,vemula2018social,sadeghian2019sophie,mangalam2020not,tang2021collaborative} and transformer structure \cite{giuliari2021transformer,yu2020spatio,yuan2021agentformer} are used to capture agents' spatial and temporal dependencies. The third is graph-based methods \cite{wu2020connecting,kosaraju2019social,huang2019stgat,chen2021human,hu2020collaborative,li2020evolvegraph,mohamed2020social,li2021online,gao2020vectornet,xu2022adaptive}, which are proposed to explicitly model the interaction between agents through non-grid structures. For example, \cite{hu2020collaborative} constructs a directed fully connected graph to model interaction between agents. \cite{li2021rain} learns a interaction graph to modeling agents relationships based on reinforcement learning. 
However, previous methods only focus on modeling the pair-wise interaction, ignoring the group behavior's influence on agents.
Some works \cite{bisagno2018group,zhou2022ga,zhou2021grouptron} attempt to model group behavior by coherent filtering, additional group annotations or clustering algorithm but they only consider a static and single-scale group for each agent.

In this work, we also want to model group behaviors by extending the graph-based mechanism from ordinary graphs to dynamic multiscale hypergraphs. 
Comparing to previous methods~\cite{bisagno2018group,zhou2022ga,zhou2021grouptron}, we are novel in i) we conduct group capturing which is learned without any manual annotation; while previous methods use non-learnable methods that directly use coordinates or use additional annotations to obtain the group; ii) we capture multi-scale groups for every agent and also consider the changing of the group belongingness through time, while previous works only capture a single-scale group for every agent and regard the group as static over time; iii) we construct hypergraph to model the interaction inner groups for multiple agents interacting together while previous methods use ordinary graph; and iv) we perform relational reasoning in the group interaction modeling while previous methods not consider.

\textbf{Relational reasoning.}
Relational reasoning is widely studied in various problems with structural dependency among data instances. Traditional approaches like locally linear embedding (LLE) \cite{roweis2000nonlinear} and Isomap \cite{tenenbaum2000global} use kNN \cite{altman1992introduction} for forming such relationships based on measures of similarities among data instances. Recently, deep learning has become
the main tool for learning these dependencies under different tasks like classification \cite{kipf2016semi}, visual reasoning~\cite{van2018relational}, visual question answering  \cite{narasimhan2018out,santoro2017simple}, meta-learning  \cite{alet2019neural} and graph representation learning  \cite{garcia2017learning}. For example, \cite{van2018relational} 
incorporates prior knowledge about the compositional nature of human perception to factor interactions between object-pairs to discover objects and model their physical interactions from
raw visual images. \cite{santoro2017simple} proposes a Relation Network that more exclusively focused on general relation reasoning and can be applied to tasks with relatively unstructured inputs.
In the multi-agent trajectory prediction task, only the trajectories of individual agents are available without knowledge of the underlying relations, thus it is challenging to explicitly infer agents' interactions and make relational reasoning.
Some works attempt to infer the interaction between agents, which are more relevant with our work. NRI \cite{kipf2018neural} infers a latent interaction graph via an autoencoder model. RAIN \cite{li2021rain} also infers an interaction graph using reinforcement learning to select edges. EvolveGraph \cite{li2020evolvegraph} and dNRI \cite{graber2020dynamic} consider an evolve mechanism and learn a dynamic interaction graph. \cite{banijamali2021neural} introduces the node-specific information to tackle the problem in a setting where agent potentially has individualized information that other agents cannot have access to.

Our work also tries on relational reasoning but the main differences are: i) we capture more interactions including both group-wise interaction and pair-wise interaction, comparing to only inferring pair-wise relationships on previous methods; ii) we infer both interaction category and interaction strength while previous works only consider the interaction category; and iii) we consider an evolving mechanism that focuses on the evolvement of whole agent's interacted embeddings with a global receptive field; while previous works only consider the evolvement of interaction category with a receptive field of only the last timestamp.

\begin{figure*}[t] 
\centering
\includegraphics[width=0.95\textwidth]{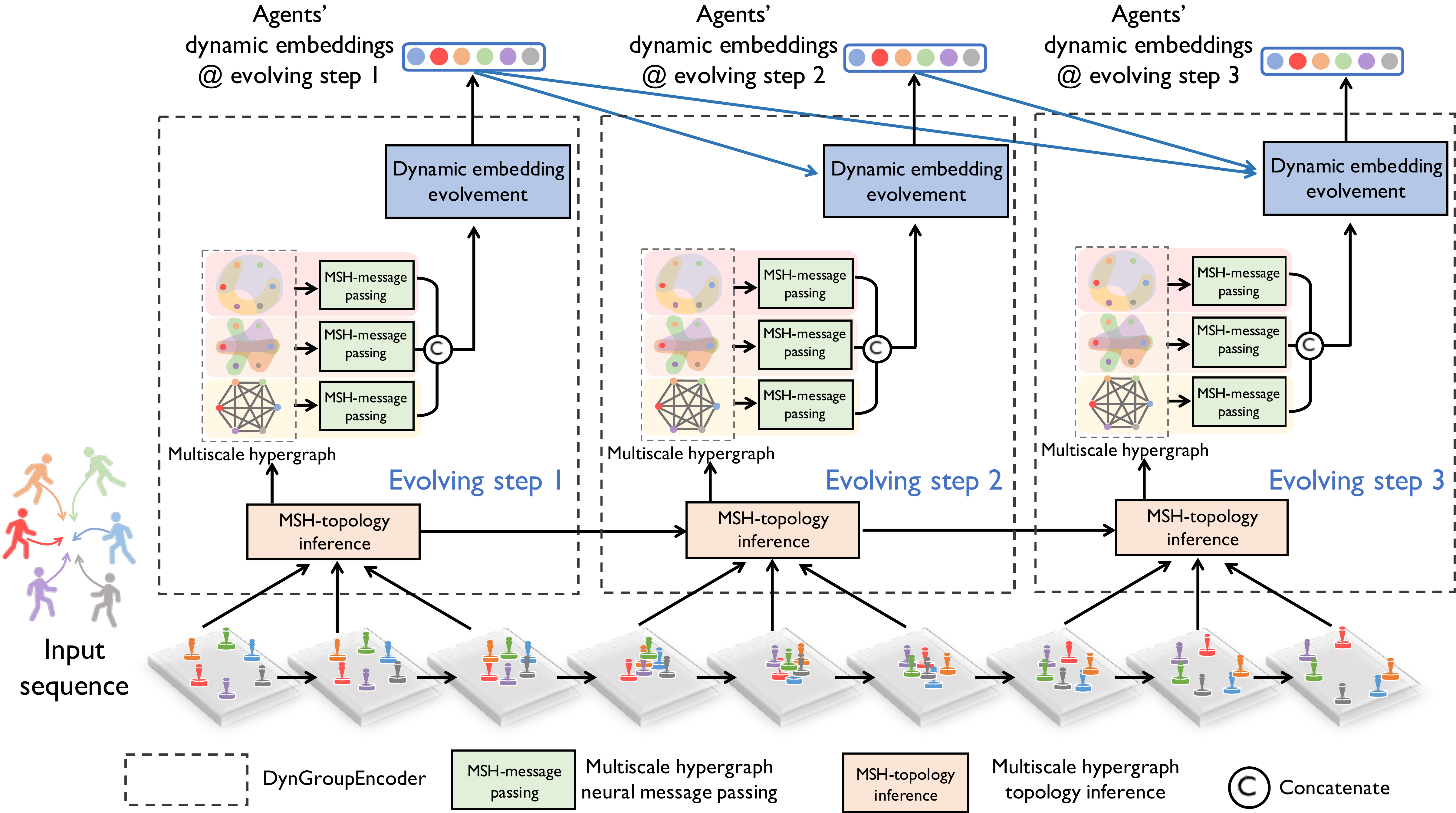}
\vspace{-2mm}
\caption{\small The architecture of~\texttt{DynGroupNet}.~\texttt{DynGroupNet} encodes agent trajectory by multiple evolving steps. The evolving step repeats by an evolving gap $\tau$ with an evolving length $T_{\rm E}$. Here we present an example of $T_{\rm E}=\tau=3$.~\texttt{DynGroupNet} uses the DynGroupEncoder (gray rectangle) on every evolving step considering both multiscale group interaction and temporal dependencies. In the DynGroupEncoder, we infer and use the dynamic multiscale hypergraph by the multiscale hypergraph topology inference and message passing to get agents' embeddings and interaction relationships. We integrate embeddings at different evolving steps by the dynamic embedding evolvement to output agents' dynamic embeddings as~\texttt{DynGroupNet}'s output. 
}
\label{fig:DynGroupNet}
\vspace{-2mm}
\end{figure*}

\section{Problem Formulation}
\label{sec:formulation}
In this paper, we study the trajectory prediction along with dynamic relational reasoning which means predicting future trajectories of multiple agents and inferring interaction relationship given their past trajectories. Mathematically, let $\mathbb{X} \in \R^{N \times (T_{\rm p}+T_{\rm f}) \times 2}$ be the whole trajectory, which consists of past trajectories
$\mathbb{X}^{-} \in \R^{N \times T_{\rm p} \times 2}$ and future trajectories $\mathbb{X}^{+} \in \R^{N \times T_{\rm f} \times 2}$ of all the $N$ agents in the scene, where $T_{\rm p}$ and $T_{\rm f}$ are the length of the past and future trajectory. 
Let $\X^{-}_i = \mathbb{X}^{-}_{i,:,:} = [\x^{0}_{i}, \x^{1}_{i},\cdots, \x^{T_{\rm p}-1}_{i}] \in \R^{T_{\rm p} \times 2}$ and $\X^{+}_i = \mathbb{X}^{+}_{i,:,:} = [\x^{T_{\rm p}}_{i}, \x^{T_{\rm p}+1}_{i},\cdots, \x^{T_{\rm p}+{T}_{\rm f}-1}_{i}] \in \R^{T_{\rm f} \times 2}$ be the past and future trajectories of the $i$th agent, respectively, where $\x_i^t \in \R^{2}$ is the 2D coordinate at timestamp $t$. 
For predicting future trajectories of multiple agents and model multimodality of possible futures, we seek to learn a diverse prediction model $g(\cdot)$, so that one of the predicted future trajectories $\widehat{\mathbb{X}}^{+} = g(\mathbb{X}^{-})$ are as close to the ground-truth future trajectories $\mathbb{X}^{+}$ as possible. 

For a more precise future prediction, the prediction model $g(\cdot)$ needs to contain agent social interaction modeling. Additionally, relational reasoning is a supplementary task in trajectory prediction
when the prediction model $g(\cdot)$ is capable to explicitly infer social influences among agents like inferring interaction category or strength. Relational reasoning can further be expanded to dynamic relational reasoning since the social influences may vary along the time. Note that only the trajectories of individual agents are available without knowledge of the underlying relations, thus the process of dynamic relational reasoning is unsupervised.

In the field of trajectory prediction, previous methods, such as \cite{alahi2016social,mangalam2020not,huang2019stgat,yu2020spatio,hu2020collaborative,li2020evolvegraph} models pair-wise interactions in the prediction model $g(\cdot)$. However, in many scenarios, group behavior is common that multiple agents interact together. In this work, we consider both pair-wise and group-wise interactions in $g(\cdot)$ through a dynamic multiscale hypergraph. Additionally, in the field of dynamic relational reasoning, previous methods, such as \cite{kipf2018neural,li2020evolvegraph} only consider the interaction category, we consider both interaction category and interaction strength in the prediction model $g(\cdot)$. In the following Section \ref{sec:dyngroupnet} and \ref{sec:system}, we will introduce the core design and the prediction system. 

% when the prediction model $g(\cdot)$ is capable to infer dynamic social influences among the agents without any ground-truth supervision about time-varying interaction category or strength since only the trajectories of individual agents are available without knowledge of the underlying relations, this supplementary task is called~\emph{unsupervised dynamic relational reasoning}. 

% let $\mathbb{X} \in \R^{N \times (T_{\rm p}+T_{\rm f}) \times 2}$ be all trajectories of $N$ agents in a scene,  Let $\X^{-}_i = \mathbb{X}^{-}_{i,:,:} = [\x^{0}_{i}, \x^{1}_{i},\cdots, \x^{T_{\rm p}-1}_{i}] \in \R^{T_{\rm p} \times 2}$ and $\X^{+}_i = \mathbb{X}^{+}_{i,:,:} = [\x^{T_{\rm p}}_{i}, \x^{T_{\rm p}+1}_{i},\cdots, \x^{T_{\rm p}+{T}_{\rm f}-1}_{i}] \in \R^{T_{\rm f} \times 2}$ be the past and future trajectories of the $i$th agent, where $\x_i^t \in \R^{2}$ is the 2D coordinate in the world space or image pixel space at time $t$. Our goal is to learn a prediction model $g(\cdot)$, so that the predicted future trajectories $\widehat{\mathbb{X}}^{+} = g( \mathbb{X}^{-})$ are as close to the ground-truth future trajectories $\mathbb{X}^{+}$ as possible. When a prediction model $g(\cdot)$ is capable to infer dynamic social influences among the agents without any ground-truth supervision about time-varying interaction category or strength, this supplementary task is called~\emph{unsupervised dynamic relational reasoning}.

\section{DynGroupNet}
\label{sec:dyngroupnet}
\subsection{Architecture overview}
The core of~\texttt{DynGroupNet} is to learn a dynamic multiscale hypergraph whose node is the agent and hyperedge is the interaction; and then, leverage this dynamic multiscale hypergraph to learn agent and interaction embeddings across time.
We consider $K$ evolving steps to capture agent dynamic interaction. In every evolving step, the input is a sub-trajectory of a time period and the output is the corresponding agent embeddings. In an evolving step, we use a DynGroupEncoder which mainly consists of three operations, that is, the multiscale hypergraph topology inference; the multiscale hypergraph neural message passing; and the dynamic embedding evolvement. The multiscale hypergraph topology inference aims to infer the topology of the multiscale hypergraph given the input sub-trajectory. The multiscale hypergraph neural message passing aims to learn the patterns of agents and their interaction relationships given the inferred multiscale hypergraph. The dynamic embedding evolvement receives the learned agents' patterns along the time and integrates them considering their temporal dependencies to get the agents' dynamic embeddings as the output. 

Mathematically, let $\mathbb{X} \in \R^{N \times (T_{\rm p}+T_{\rm f}) \times 2}$ be the whole input trajectory of \texttt{DynGroupNet}.
At the $k$th evolving step, the input sub-trajectory is $\mathbb{X}^{[k]} = \mathbb{X}_{:,k\tau:k\tau+T_{\rm E}-1,:} \in \R^{N \times T_{\rm E} \times 2}$, which means we perform an evolving step by every evolving gap $\tau$ and take a input sub-trajectory of evolving length $T_{\rm E}$. Fig. \ref{fig:DynGroupNet} shows the overview \texttt{DynGroupNet} architecture where we set $T_{\rm E} = \tau =3$. 

\subsection{Multiscale hypergraph topology inference}
\label{subsection:topology}
\begin{figure}[t] 
\centering
\includegraphics[width=0.43\textwidth]{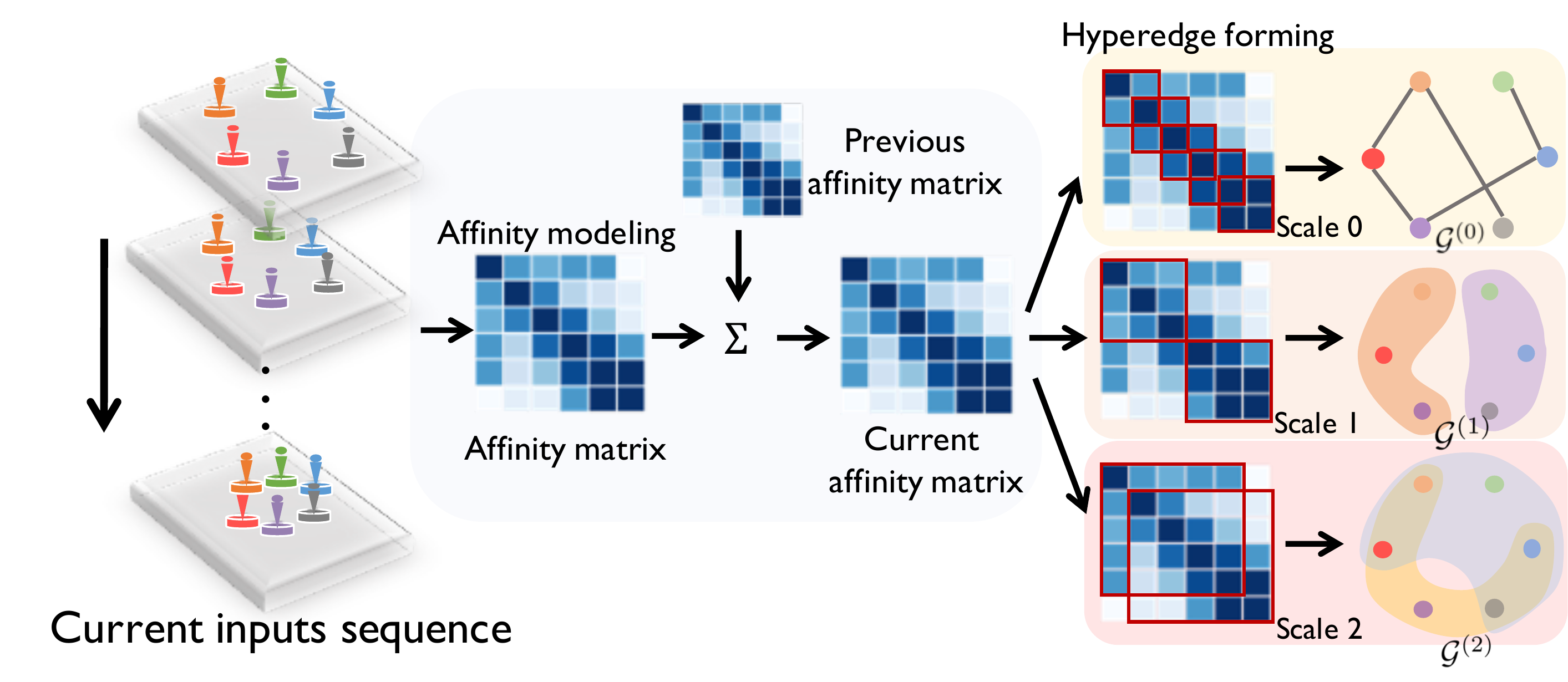}
\caption{\small The procure of multiscale hypergraph topology inference. We calculate the affinity matrix infer the hypergraph topology at different scales from the affinity matrix.}
\vspace{-2mm}
\label{fig:topology}
\end{figure}

Given the $k$th evolving step input $\mathbb{X}^{[k]}$, to comprehensively model its group-wise interactions at multiple scales, we consider inferring a multiscale hypergraph from agent dynamics to reflect interactions at various group sizes, see Fig~\ref{fig:topology}. Mathematically, let $\mathcal{V} = \{ v_1, v_2, \cdots, v_N \}$ be a set of agents and $\mathcal{G}^{[k]} = \{\mathcal{G}^{(0,k)}, \mathcal{G}^{(1,k)}, ... \mathcal{G}^{(S,k)} \}$ be a multiscale hypergraph showing the agent connections at the $k$th evolving step, where $\mathcal{G}^{(s,k)} =(\mathcal{V}, \mathcal{E}^{(s,k)})$ denotes the hypergraph of the $s$th scale and the $k$th evolving step.  The hyperedge set $\mathcal{E}^{(s,k)} = \{e^{(s,k)}_1, e^{(s,k)}_2, \cdots, e^{(s,k)}_{M_s} \}$ represents group-wise relations with $M_s$ hyperedges, each of which links a number of agents to represent the common relations. A larger $s$ indicates a larger scale of agent groups. Notably, $\mathcal{G}^{(0,k)} =(\mathcal{V},\mathcal{E}^{(0,k)})$ is a specific hypergraph whose edges model the finest pair-wise agent connections. The topology of each $\mathcal{G}^{(s,k)}$ can be represented as an incidence matrix $\mathbf{H}^{(s,k)} \in \R^{|\mathcal{V}|\times |\mathcal{E}^{(s,k)}|}$ where $\mathbf{H}^{(s,k)}_{i, j}=1$ if the $i$th node is included in the $j$th hyperedge, otherwise $\mathbf{H}^{(s,k)}_{i, j}=0$.

\textbf{Affinity modeling.} The trajectory interactions are in a stealth mode and nontrivial to capture since there are no explicitly predefined graph topology for relation description. To infer a multiscale hypergraph at the $k$th evolving step, we construct hyperedges by grouping agents that have highly correlated trajectories, whose correlations could be measured by mapping the trajectories as a high-dimensional feature vector. Concretely, given the input trajectory of $i$th agent $\X^{[k]}_i$ at the $k$th evolving step, let $\mathbf{q}_i^{[k]} = f_{\mathrm{init}}({\X^{[k]}_i}) \in \R^{d},$ be its initial trajectory embedding, where $f_{\mathrm{init}}(\cdot)$ can be a MLP. We then compute an affinity matrix $\mathbf{A}^{[k]} \in \R^{N\times N}$ to reflect the correlation of any two agents at the $k$th evolving step. The $(i,j)$th element of $\mathbf{A}^{[k]}$ is:
\begin{equation*}
\label{eq:attention}
    \mathbf{A}^{[k]}_{i,j} \ = \  \mathbf{q}_i^{[k]^{\top}} \mathbf{q}^{[k]}_j / (\|\mathbf{q}_i^{[k]}\|_2 \|\mathbf{q}^{[k]}_j\|_2),
\end{equation*}
which represents the relational weight of the $i$th agent and the $j$th agent reflecting the correlation between two agents at the $k$th evolving step. We normalize the embeddings to ensure the trajectory representations to carry unit energy for stable relation estimation.

To maintain and propagate the history group behavior for a smoothing varying of agents affinity, we adjust the current $k$th $(k>0)$ affinity matrix $\mathbf{A}^{[k]}$ by a weight sum of the last affinity matrix $\mathbf{A}^{[k-1]}$. Mathematically, the adjustment is:
\begin{equation*}
\mathbf{A}^{[k]} = (\mathbf{A}^{[k]}+\alpha \mathbf{A}^{[k-1]})/(1+\alpha),
\end{equation*}
where $\alpha \in [0,1]$ is a hyperparameter adjusting the weight of the previous affinity matrix.

\textbf{Hyperedge forming.} Given the affinity matrix $\mathbf{A}^{[k]}$ at the $k$th evolving step, we then form hyperedges at various scales. At the $0$th scale, we consider pair-wise connections. Each node connects the nodes that have the largest affinity scores with it, leading to $M_0$ edges in the incident matrix $\mathbf{H}^{(0,k)}$. For the other scales, we consider group-wise connections. Intuitively, agents in a group should have high correlation with each other. We thus find the groups of nodes by looking for high-density submatrices in the affinity matrix $\mathbf{A}^{[k]}$. We first assign a sequence of increasing group sizes $\{M^{(s)}\}_{s=1}^S$; and then, for each node, we find a highly correlated group at any scale $s$, leading to $N$ groups/hyperedges in each scale. 

Mathematically, let the $i$th hyperedge $e^{(s,k)}_i$ at the $s$th scale at the $k$th evolving step be the one associated with the $i$th node $v_i$. This hyperedge is obtained by solving the following optimization problem of searching a $M^{(s)}\times M^{(s)}$ submatrix:
\begin{equation}
   \setlength{\abovedisplayskip}{2pt}
   \setlength{\belowdisplayskip}{1pt}
\label{eq:opt}
   \begin{aligned}
   & e^{(s,k)}_i = \arg \mathop{\operatorname{max}}\limits_{\Omega^{(s,k)} \subseteq \mathcal{V}} \left \| \mathbf{A}^{[k]}_{\Omega^{(s,k)}, \Omega^{(s,k)}}\right\|_{\mathrm 1,\mathrm 1}, \\ 
   & \mathrm{s.t.}\;|\Omega^{(s,k)}|=M^{(s)}, ~v_i \in \Omega^{(s,k)}, ~ i=1,\dots,N,
   \end{aligned}
\end{equation}
where the entry-wise matrix norm $\|\cdot\|_{\mathrm 1,\mathrm 1}$ denotes the sum of the absolute values of all elements. The objective finds the most correlated agents and links them together to consider the group behaviors. The first constraint limits the group size and the second requires the participance of the $i$th node. In this way, we could form at least one hyperedge that belongs to each node at each scale. To solve the optimization~\eqref{eq:opt}, when the total number of agents is small, we could adopt an enumeration algorithm to search for the optimum solution; otherwise, we employ a greedy algorithm approximation that first selects $v_i$ and then adds new nodes of maximum affinity value with $v_i$ at each move sequentially.

The incidence matrices for a multiscale hypergraph of the $k$th evolving step are thus $\{\mathbf{H}^{(0,k)} \in \R ^{N\times NM^{(0)}}, \{ \mathbf{H}^{(s,k)} \in \R ^{N\times N} \}_{s=1}^S\}$, where $\mathbf{H}^{(0,k)}$ includes $NM^{(0)}$ edges to consider the pair-wise interactions and $\mathbf{H}^{(s,k)} (s \geq 1)$ includes $N$ hyperedges to reflect the group-wise interactions within $M^{(s)}$ agents. Different from the common multiscale graph whose node numbers vary from scales \cite{li2020dynamic,gao2019graph,li2021multiscale}, our multiscale hypergraph has fixed node numbers, yet different hyperedge sizes on various scales. 
Note that all the edges and hyperedges are selected based on the same affinity matrix. This design brings two benefits: i) it is computationally efficient to search for high-order relationships from a single matrix; and ii) it makes the training of the affinity matrix more stable and informative through back-propagation.

Compared to~\cite{kipf2018neural,hu2020collaborative,li2020evolvegraph}, our topology inference method is novel from two aspects. First, we actively infers a graph structure through learning; while many previous methods directly adopt a fixed fully-connected topology. Second, our method models both pair-wise and group-wise connections at multiple scales; while the previous methods only model the pair-wise connections at a single scale.

\subsection{Multiscale hypergraph neural message passing}
\begin{figure}[t] 
\centering
\includegraphics[width=0.48\textwidth]{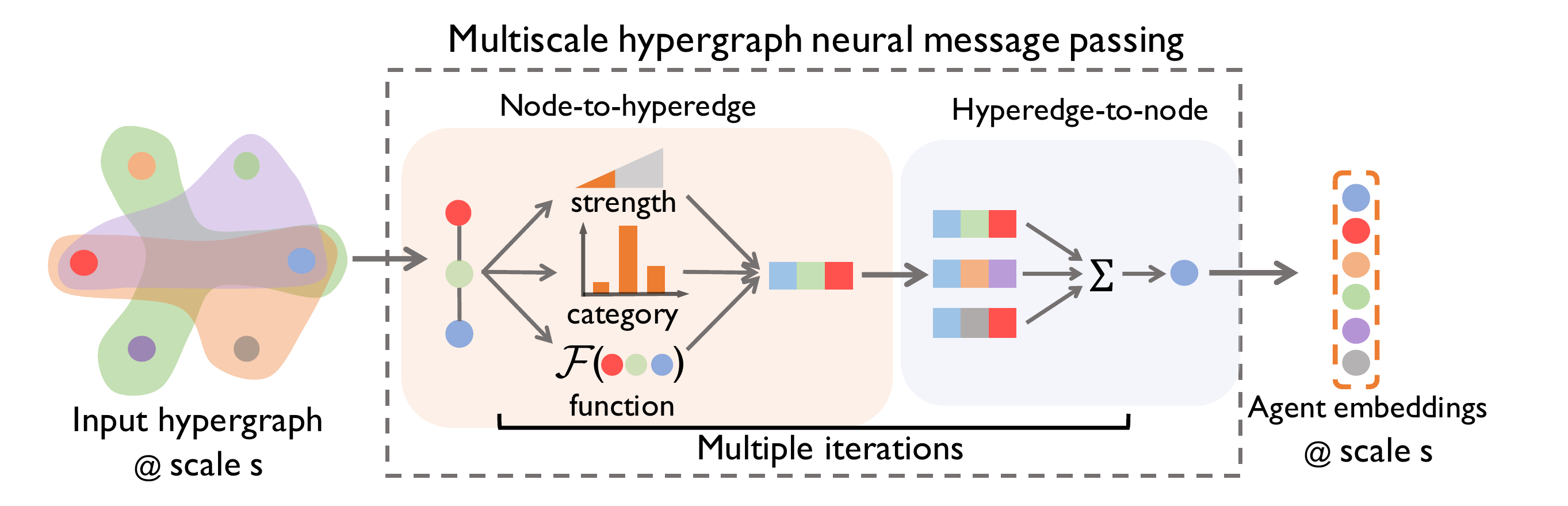}
\caption{\small The procure of multiscale hypergraph neural message passing. In the node-to-hyperedge phase, we using related agents' embeddings to calculate their interaction embedding with three elements: neural interaction strength, neural interaction category and per-category function. In the hyperedge-to-node phase, we update agent embedding by considering all related interactions. } 
\label{fig:ms-nmp}
\vspace{-2mm}
\end{figure}

To learn the patterns of agents trajectories of the $k$th evolving step given the inferred multiscale hypergraph, we customize a multiscale hypergraph neural message passing method obtaining the embeddings of agents and interactions iteratively through node-to-hyperedge and hyperedge-to-node, see Fig \ref{fig:ms-nmp}.
Specifically, we first initialize the agent embedding from trajectory of each agent; that is, at scale $s$ of evolving step $k$, for the $i$th agent, $v_i$, its initial embedding is $\mathbf{v}_{i}^{(s,k)} = \mathbf{q}^{[k]}_i \in \mathbb{R}^d$ (see Sec.~\ref{subsection:topology}). At each scale, in the node-to-hyperedge phase, groups of agent embeddings are aggregated to get the interaction embeddings. In the hyperedge-to-node phase, each agent embedding is updated according to associated interaction embeddings. We execute the node-to-hyperedge and hyperedge-to-node for several iterations. We finally obtain the representation of each agent by fusing its embeddings across all the scales.

\textbf{Node-to-hyperedge phase.} To promote a representation for relational reasoning, we exhibit an interaction embedding with three elements: \textit{neural interaction strength}, representing the intensity of the interaction, \textit{neural interaction category}, reflecting agents interaction category, and \textit{per-category function}, modeling how interaction process of this category works. At the $s$th scale of hypergraph of evolving step $k$, for the $i$th hyperedge, $e_{i}^{(s,k)}$, its interaction embedding is obtained as:
\begin{equation}
\label{eq:decouple}
   \setlength{\abovedisplayskip}{1pt}
   \setlength{\belowdisplayskip}{1pt}
    \mathbf{e}_{i}^{(s,k)} = r_{i}^{(s,k)}  \sum\limits_{\ell=1}^L  c_{i,\ell}^{(s,k)} \mathcal{F}^{(s)}_{\ell} \Big(\sum\limits_{v_j \in e_{i}^{(s,k)}} \mathbf{v}_{j} \Big) \in \R^{d}, 
\end{equation}
where $r_{i}^{(s,k)}$ is its neural interaction strength. $c_{i,\ell}^{(s,k)}$, the $\ell$th element of a category vector ${\bf c}_{i}^{(s,k)} \in [0,1]^{L}$, denotes the probability of the $\ell$th neural interaction category within $L$ possible categories. For each category at scale $s$, we assign a learnable category function $\mathcal{F}_{\ell}^{(s)}(\cdot)$ implemented by MLPs. Each of these three elements is trainable in an end-to-end framework.

To obtain the neural interaction strength $r_{i}^{(s,k)}$, and neural interaction category ${\bf c}_{i}^{(s,k)}$ at the $s$th scale and the $k$th evolving step, we leverage a collective embedding that is a hidden state to reflect the overall information of agents in a group. For hypergraph at the $s$th scale, the collective embedding $\mathbf{z}_{i}^{(s,k)}$ of $e_{i}^{(s,k)}$ is obtained by the weighted sum of all the agent embedding associated with the hyperedge $e_{i}^{(s,k)}$; that is,
$\setlength{\abovedisplayskip}{4pt}
   \setlength{\belowdisplayskip}{4pt}
    \mathbf{z}_{i}^{(s,k)} = \sum\limits_{v_j\in e_{i}^{(s,k)}} w_{j}^{(s,k)} \mathbf{v}_{j}$, 
where $w_{j}^{(s,k)} = \mathcal{F}^{(s)}_{\rm w} \Big( \mathbf{v}_{j}, \sum\limits_{v_m \in e_{i}^{(s,k)}} \mathbf{v}_{m} \Big),$
with $\mathcal{F}^{(s)}_{\rm w}(\cdot)$ implemented by a MLP. The weight $w_{j}^{(s,k)}$ reflects the contribution of the $j$th node to the $i$th group at scale $s$ at the $k$th evolving step. We then use the collective embedding to infer the neural interaction strength, $r_{i}^{(s,k)}$, and the neural interaction category, ${\bf c}_{i}^{(s,k)}$; that is,
\begin{equation*}
\begin{aligned}
    r_{i}^{(s,k)} &= \sigma\left(\mathcal{F}^{(s)}_{\rm r}(\mathbf{z}_{i}^{(s,k)})\right), \\
\mathbf{c}_{i}^{(s,k)}  &=   \mathrm{softmax} \left(( \mathcal{F}^{(s)}_{\rm c}( \mathbf{z}_{i}^{(s,k)})+\mathbf{g})/{\gamma}\right),
\end{aligned}
\end{equation*}
where $\sigma(\cdot)$ is a sigmoid function to constrain the strength values, $\mathbf{g}$ is a vector whose elements are i.i.d. sampled from $\mathrm{Gumbel}(0,1)$ distribution and $\gamma$ is the temperature controlling the smoothness of type distribution. We use the Gumbel softmax to make a continuous approximation of the discrete distribution following~\cite{maddison2016concrete}. Functions $\mathcal{F}^{(s)}_{\rm r}(\cdot)$ and $\mathcal{F}^{(s)}_{\rm c}(\cdot)$ are used to calculate the neural interaction strength and category at the $s$th scale, which are modeled by MLPs. The inference of the neural interaction strength, neural interaction category and per-category function considers the group behavior through the shared collective embedding.

\textbf{Hyperedge-to-node phase.} Given the interaction embeddings learnt from groups of agents, we then update each agent embedding by considering all the associated interactions. Let $\mathcal{E}_i^{(s,k)} = \{e_{j}|v_{i} \in e_{j}, e_{j} \in \mathcal{E}^{(s,k)}\}$ be the set of hyperedges associated with the $i$th node $v_i$ in the $s$th scale of hypergraph at the $k$th evolving step. The embedding of the $i$th agent in the $s$th scale at the $k$th evolving step is updated as 
\begin{equation*}
     \setlength{\abovedisplayskip}{2pt}
   \setlength{\belowdisplayskip}{1pt}
\mathbf{v}_{i}^{(s,k)} \leftarrow f^{(s)}_{\rm v} \Big(\big[\mathbf{v}_{i}^{(s,k)}, \sum\limits_{e_j \in \mathcal{E}_i^{(s,k)}} \mathbf{e}_{j} \big]\Big) \in \mathbb{R}^d,
\end{equation*}
where $f^{(s)}_{\rm v}(\cdot)$ is a trainable MLP; $[\cdot ,\cdot]$ denotes the embedding concatenation of one node and the associated hyperedges to absorb the influence from interactions to an agent.

At any scale $s$ at the $k$th evolving step, we repeat the node-to-hyperedge and hyperedge-to-node phases for several times and obtain the embedding of the $i$th node, $\mathbf{v}_{i}^{(s,k)}$.
In this way, we compute the embeddings of the $i$th node, $\mathbf{v}_{i}^{(0,k)}, \mathbf{v}_{i}^{(1,k)} \dots, \mathbf{v}_{i}^{(S,k)}$, across $S$ scales in parallel. We finally concatenate the agent embedding across all the scales to obtain the $i$th agent's embedding at the $k$th evolving step.

\begin{equation}
\label{eq:embedding}
\mathbf{v}^{[k]}_{i} = [\mathbf{v}^{(0,k)}_{i}, \mathbf{v}^{(1,k)}_{i}, \cdots, \mathbf{v}^{(S,k)}_{i}] \in \mathbb{R}^{d(S+1)}.
\end{equation}

Compared to previous methods \cite{kipf2018neural},~\cite{graber2020dynamic}, and \cite{li2020evolvegraph}, which achieve relational reasoning using neural message passing, our hypergraph neural message passing is novel from three aspects. First, \cite{kipf2018neural,graber2020dynamic,li2020evolvegraph} adopt neural message passing on ordinary graphs; while we consider neural message passing on a series of hypergraphs, promoting more comprehensive information propagation and aggregation. Second, ~\cite{kipf2018neural,graber2020dynamic,li2020evolvegraph} only infer the interaction category; while our method can infer both interaction category and strength through a three-element representation format. Third, the interaction category inferred by \cite{kipf2018neural,graber2020dynamic,li2020evolvegraph} is the final output after message passing; while the interaction category and strength inferred by our method are the intermediate features that involve in neural message passing. Our design not only promotes relational reasoning in neural message passing, but also improves the prediction performance.

\subsection{Dynamic embedding evolvement}
Now an agent's embedding at each evolving step is localized at a limited range of time window. To promote the temporal dependencies of an agent's embedding across various evolving steps, we design a dynamic embedding evolvement mechanism; that is, for each agent, we upgrade its embedding in~\eqref{eq:embedding} to a dynamic embedding by aggregating its historical information at each previous evolving step. To implement this evolvement mechanism, we use a transformer structure to capture global temporal dependencies, where the query is the current embedding and the key \& value is the previous dynamic embeddings. 

\begin{figure}[t] 
\centering
\includegraphics[width=0.45\textwidth]{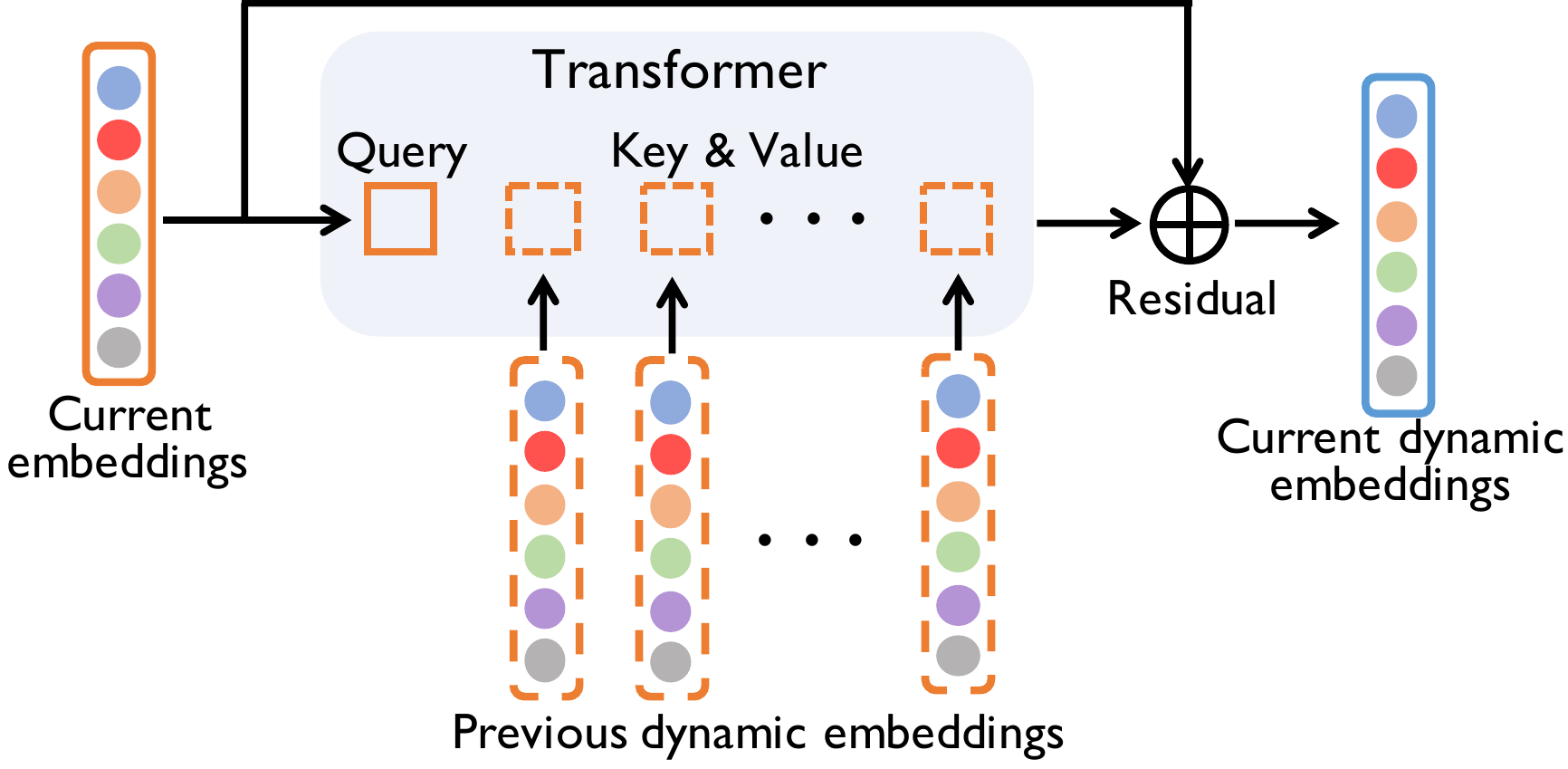}
\vspace{-2mm}
\caption{\small The procure of dynamic embedding evolvement. The current and previous embeddings are treated as query and key/value in a transformer. }
\label{fig:evolvement}
\vspace{-3mm}
\end{figure}

Let $\widetilde{\mathbf{V}}_i^{[-k]} = [\widetilde{\mathbf{v}}_i^{[0]}+p^{[0]},\cdots,\widetilde{\mathbf{v}}_i^{[k-1]}+p^{[k-1]}] \in \R^{d(S+1) \times k}$ be all the dynamic embeddings of the $i$th agent before the $k$th evolving step, where $\widetilde{\mathbf{v}}_i^{[k']}$ is the $i$th agent's dynamic embedding at the $k'$th evolving steps ($k'\in \{0,1,...,k-1\}$), $p^{[k]} \in \R^{d(S+1) \times k}$ is a positional vector and $[\cdot,\cdot]$ represents concatenation. Note that, to inform the transformer about the temporal order, we apply a positional encoding to each evolving step. The positional vector is obtained as $p^{[k']}=\mathcal{F}_{\rm pe}(-1+2\frac{k'}{k})$ with $\mathcal{F}_{\rm pe}(\cdot)$ a MLP function. Then, at the $k$th evolving step, the dynamic embedding of the $i$th agent is obtained as:
\begin{equation*}
     \widetilde{\mathbf{v}}_i^{[k]} =  \mathbf{v}_i^{[k]} + \mathrm{MHA}(\mathbf{v}_i^{[k]},\widetilde{\mathbf{V}}_i^{[-k]},\widetilde{\mathbf{V}}_i^{[-k]}) \in \R^{d(S+1)},
\end{equation*}
where $\mathrm{MHA}(\cdot)$ is a multi-head attention.  The procedure of dynamic embedding evolvement is shown in Fig.~\ref{fig:evolvement}.

Dynamic embedding evolvement significantly distinguishes~\texttt{DynGroupNet} from~\cite{xu2022groupnet}. In \cite{xu2022groupnet}, the embedding is static since only a static multiscale hypergraph is used. In~\texttt{DynGroupNet}, we consider a dynamic multiscale hypergraph and dynamic embedding evolvement is used to model the temporal relationship to aggregate interaction information of previous timestamps. Compared to other previous methods~\cite{graber2020dynamic,li2020evolvegraph} which also models the evolution of the agents' state, our dynamic embedding evolvement has two advantages. First, previous methods adopt recurrent structure like LSTM or GRU which only consider the effect of the last evolving step; while our dynamic embedding evolvement utilizes a transformer structure that has a global receptive field for all previous evolving steps, promoting more long-range temporal dependencies. Second, previous methods only focus on the evolvement of interaction category; while our method focuses on the evolvement of whole agents interacted embeddings not only includes the interaction category information, but also includes multiscale group behavior and interaction strength information, which promotes a more comprehensive evolvement of agent states.

\begin{figure*}[t] 
\centering
\includegraphics[width=0.9\textwidth]{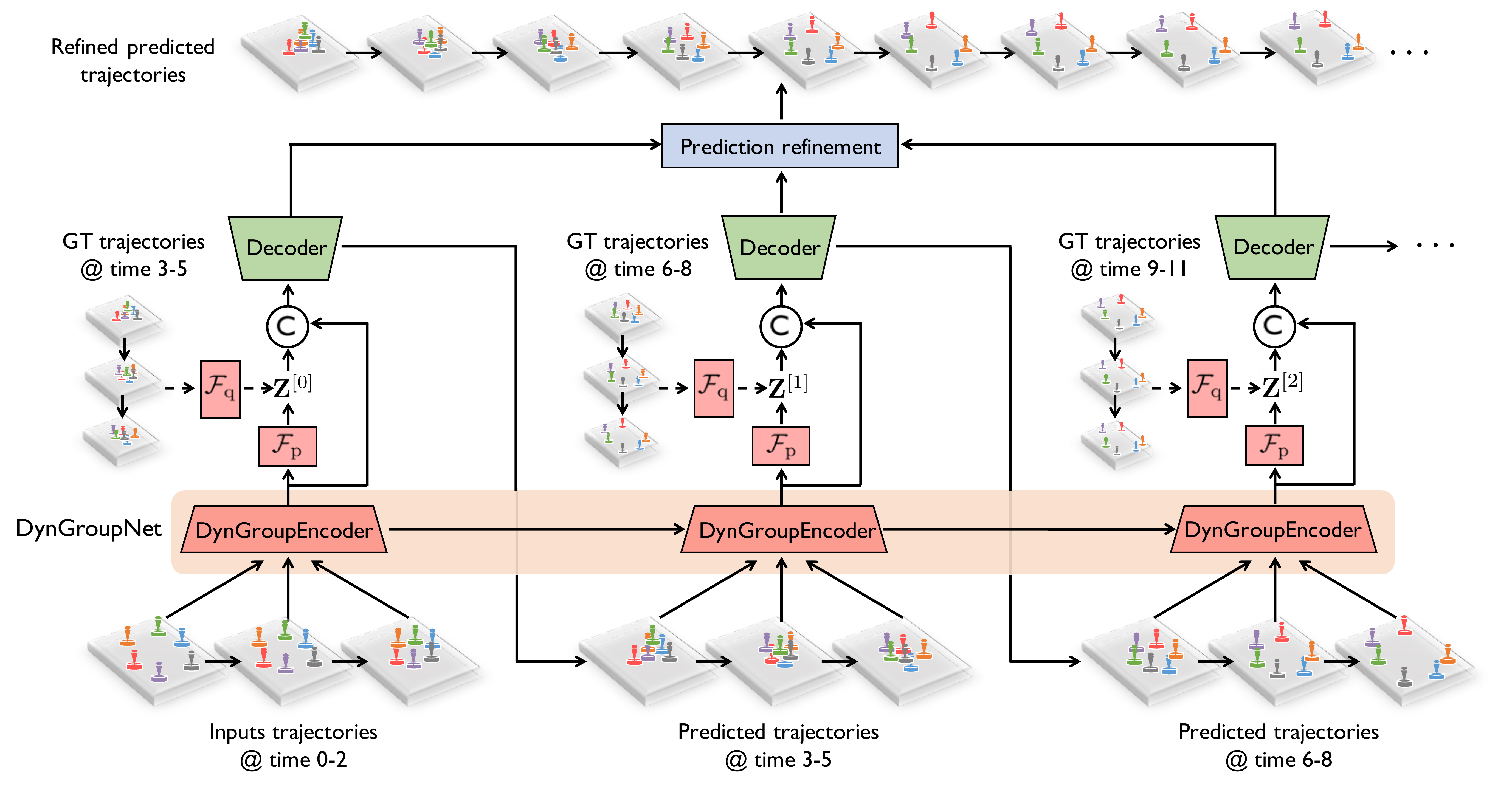}
\vspace{-2mm}
\caption{\small The architecture of the prediction system. The prediction system contains a~\texttt{DynGroupNet}-based encoding process (red), a recurrent decoding process (green), and a prediction refinement process (blue). The encoding process generates the latent codes and decoding process output the trajectories using the latent codes. Dash lines represent only use in training. The encoding-decoding process will recurrently move forward by setting output trajectories to next encoding inputs. Here we present a sample of encoding length $T_{\rm E}=3$ and decoding length $T_{\rm D}=3$. After predicting trajectories of all future timestamps, we collect the prediction and execute a prediction refinement operation to obtain final prediction results. }
\label{fig:framework}
\vspace{-3mm}
\end{figure*}

\section{Prediction System}
\label{sec:system}
Here we introduce a prediction system that cooperated
with the proposed DynGroupNet as the core module 
and how it predict dynamically-feasible future trajectories recurrently. 

The prediction system is mainly based on conditional variational autoEncoder (CVAE) structure to handle the stochasticity of each agent’s behavior. The prediction system contains a~\texttt{DynGroupNet}-based encoding process, a recurrent decoding process, and a prediction refinement process; see Fig. \ref{fig:framework}. Given the past or the predicted trajectories, the encoding process aims to generate latent codes representation through calculating agents' dynamic embeddings by~\texttt{DynGroupNet}. The decoding process aims to generate a period of agents' future trajectories given the latent codes and agents' dynamic embeddings by the decoder. The encoding-decoding process will repeat recurrently by treating the predicted trajectories as part of the next input trajectories. Finally, the prediction refinement process will collect the predicted trajectories across all encoding-decoding processes and refine them together to output the refined predicted trajectories as the final output of our prediction system.

To fully use the observed past trajectories $\mathbb{X}^{-} \in \R^{N\times T_{\rm p}\times 2}$, we set the evolving length $T_{\rm E} = T_{\rm p}$. To make a non-overlap prediction, we set the length of predicted trajectories by the decoder in one decoding process equals to the evolving gap $T_{\rm D} = \tau$. Fig. \ref{fig:framework} gives an example of $T_{\rm E} = T_{\rm p} = 3$, $T_{\rm D} = \tau = 3$. To simplify following equations, let $\widehat{\mathbb{X}}^{[k]}_{\rm in}=[\widehat{\mathbf{X}}^{kT_{\rm D}},\cdots,\widehat{\mathbf{X}}^{kT_{\rm D} +T_{\rm E}-1}] \in \R^{N\times T_{\rm E} \times 2}$ and $\mathbb{X}^{[k]}_{\rm out} = [\mathbf{X}^{kT_{\rm D}+T_{\rm E }},...,\mathbf{X}^{(k+1)T_{\rm D}+T_{\rm E}-1}]\in \R^{N\times T_{\rm D} \times 2}$ be the input and the ground-truth output of the $k$th encoding-decoding process, where the superscript $[k]$ denotes the $k$th encoding-decoding process and $~\widehat{}~$ denotes the sequence is predicted. Note that for historical timestamps $t<T_{\rm p}$ we set $\widehat{\mathbf{X}}^{t}=\mathbf{X}^{t}$ since we have the historical sequence.

\subsection{Encoding process}
Given the inputs of $k$th encoding process $\widehat{\mathbb{X}}^{[k]}_{\rm in} \in \R^{N \times T_{\rm E} \times 2}$ and its corresponding ground-truth $\mathbb{X}^{[k]}_{\rm out} \in \R^{N\times T_{\rm D} \times 2}$, the encoding process first generates the parameters of the latent codes' distribution and then sample the latent code representation $\mathbf{Z}^{[k]}\in \R^{N\times d_z}$ from the distribution, $d_z$ represent the dimension of the every agent's latent codes. To model the multi-modality of the future trajectories, we introduce a overall categorical distribution ${\rm Cat}(d_z,\bm{\mu})$ for all $N$ agents, where the distribution parameters of $i$th agent $\mu_i$ represent a categorical distribution, satisfying $\|\bm{\mu}_i\|_1=1, \bm{\mu}_{i,j} \in [0,1], j \in \{0,1,\cdots,d_z-1\}$. We then sample the one-hot discrete categorical latent codes' representation of every agent from its distribution. Mathematically, the distribution parameter generation is given by:
\begin{subequations}
\label{eq:ecd}
\begin{align}
    &\widetilde{\mathbf{V}}^{[k]} = {\rm DGN}(\widehat{\mathbb{X}}^{[k]}_{\rm in}), \;\mathbf{V}^{[k]}_{\rm gt} = {\mathcal{F}_{\rm f}}(\mathbb{X}^{[k]}_{\rm out}), \label{eq:ecd1}\\
    &\bm{\mu}_{\rm p}^{[k]}= \mathcal{F}_{\rm p}(\widetilde{\mathbf{V}}^{[k]}), \; \bm{\mu}_{\rm q}^{[k]}= \mathcal{F}_{\rm q}\left([\widetilde{\mathbf{V}}^{[k]},\mathbf{V}^{[k]}_{\rm gt}]\right),\label{eq:ecd2}\\
    &p(\mathbf{Z}^{[k]}|\widehat{\mathbb{X}}^{[k]}_{\rm in})) = {\rm Cat}(d_z,\bm{\mu}_{\rm p}^{[k]}), \label{eq:ecd3}\\
    &q(\mathbf{Z}^{[k]}|\widehat{\mathbb{X}}^{[k]}_{\rm in},
    \mathbb{X}^{[k]}_{\rm out}) = {\rm Cat}(d_z,\bm{\mu}_{\rm q}^{[k]}).\label{eq:ecd4}
\end{align}
\end{subequations}
% \Note{sc: this array of equations is long, i would give each equation an index, like (a), (b), (c). Then, in the text, i would say: in step (a)， we computes ***; in step (b), ***， more organized}
In (\ref{eq:ecd1}), we obtain the current agents dynamic embeddings using proposed ~\texttt{DynGroupNet} ${\rm DGN}(\cdot)$ and future agents embeddings using a future trajectories encoding module $\mathcal{F}_{\rm f}(\cdot)$, which can be implemented by LSTM or MLP. In (\ref{eq:ecd2}), we calculate the parameters of prior and posterior distribution of $k$th encoding process $\bm{\mu}_{\rm p}^{[k]}$ and $\bm{\mu}_{\rm q}^{[k]}$ through linear layers $\mathcal{F}_{\rm p}(\cdot)$ and $\mathcal{F}_{\rm q}(\cdot)$, respectively. In (\ref{eq:ecd3}) and (\ref{eq:ecd4}), we obtain the prior distribution $p(\mathbf{Z}^{[k]}|\widehat{\mathbb{X}}^{[k]}_{\rm in})$ and the posterior distribution $q(\mathbf{Z}^{[k]}|\widehat{\mathbb{X}}^{[k]}_{\rm in},
\mathbb{X}^{[k]}_{\rm out})$. We sample latent code of possible future trajectories $\mathbf{Z}^{[k]} \sim  q(\mathbf{Z}^{[k]}|\widehat{\mathbb{X}}^{[k]}_{\rm in},
\mathbb{X}^{[k]}_{\rm out})$ in the training phase. In the testing phase, since the ground-truth trajectories are not available, we sample
$\mathbf{Z}^{[k]}$ from the prior distribution $\mathbf{Z}^{[k]}  \sim  p(\mathbf{Z}^{[k]}|\widehat{\mathbb{X}}^{[k]}_{\rm in})$. We concatenate the latent code $\mathbf{Z}^{[k]}$ with the agents dynamic embedding $\widetilde{\mathbf{V}}^{[k]}$ as the output of encoding process: $\mathbf{V}_{\mathrm{E}}^{[k]} =[\mathbf{Z}^{[k]},\widetilde{\mathbf{V}}^{[k]}]$. 

The design intuition of the encoding process is to handle the stochasticity of each agent’s behavior. We sample the latent code from its distribution to represent agent multimodel behavior information. We concatenate the latent code with the current agents' dynamic embeddings to both consider potential multimodel behavior information and current movement information.

\subsection{Decoding process}
\begin{figure}[t] 
\centering
\includegraphics[width=0.45\textwidth]{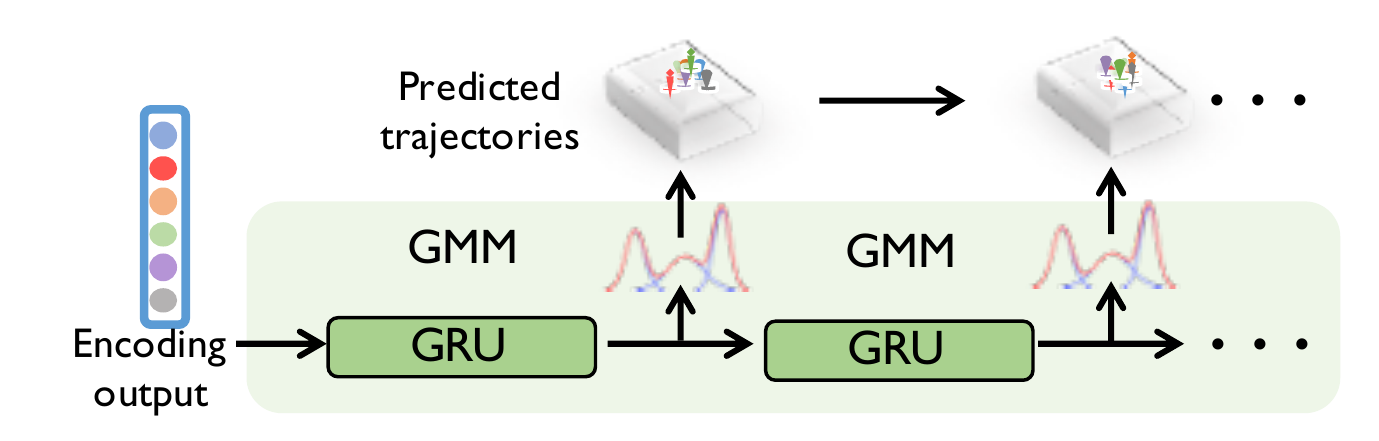}
\caption{\small The architecture of the prediction decoder. We use GMM to model the uncertainty of future trajectories and use a GRU to recurrently regress the distribution parameters of GMM.}
\label{fig:decoder}
\vspace{-2mm}
\end{figure}

After obtaining the output of encoding process $\mathbf{V}^{[k]}_{\mathrm{E}}$, we apply a recurrent decoder to obtain the prediction of $k$th encoding-decoding process $\widehat{\mathbb{X}}^{[k]}_{\rm out} = [\widehat{\mathbf{X}}^{kT_{\rm D}+T_{\rm E}},...,\widehat{\mathbf{X}}^{(k+1)T_{\rm D}+T_{\rm E}-1}]$, see Fig. \ref{fig:decoder}. To better model the uncertainty of the predicted trajectories, we calculate the distribution 
of the predicted trajectories instead of directly regress the trajectory coordinates deterministically. To model the predicted trajectories' distribution, here we adopt a bivariate Gaussian distribution model. 

The recurrent decoder consists of a gated recurrent unit (GRU) and a bivariate Gaussian distribution model. Each GRU cell generates the parameters of the bivariate Gaussian distribution step by step and the bivariate Gaussian distribution outputs agents' next velocity through sampling. Mathematically, in the $k$th decoding process, the $l$th generation step ($l \in \{0,...,T_{{\rm D}}-1\}$) is formulated as:
\begin{subequations}
    \begin{align}
        &\mathbf{M}^{[k],l} = {\mathrm{ GRU}} \left(\mathbf{M}^{[k],l-1},~ [\mathbf{V}^{[k]}_{\mathrm{E}},\widehat{\mathbf{X}}^{kT_{\rm D}+T_{\rm E}+l-1}] \right), \label{eq:dec1}\\
        & \widehat{\Delta \mathbf{X}}^{kT_{\rm D}+T_{\rm E}+l} \sim \mathrm{GMM} \left(\mathcal{F}_{\rm GMM}(\mathbf{M}^{[k],l})  \right), \label{eq:dec2}\\
        &\widehat{\mathbf{X}}^{kT_{\rm D}+T_{\rm E}+l}= \widehat{\mathbf{X}}^{kT_{\rm D}+T_{\rm E}+l-1} +   \widehat{\Delta \mathbf{X}}^{kT_{\rm D}+T_{\rm E}+l}.\label{eq:dec3}
    \end{align}
\end{subequations}
In (\ref{eq:dec1}), we use a GRU cell to recurrently calculate the hidden states, where $\mathrm{GRU}(\cdot)$ represents a GRU cell and $\mathbf{M}^{[k],l}$ denotes the hidden states of GRU in the $l$th step. In (\ref{eq:dec2}), we generate distribution parameters of GMM using a linear layer $\mathcal{F}_{\rm GMM}(\cdot)$. We then sample the velocity from the GMM distribution $\mathrm{GMM}(\cdot)$. In (\ref{eq:dec3}), we add the velocity and obtain the coordinates of next timestamps. 

We obtain the predicted trajectories of $k$th encoding-decoding process by execute $T_{\rm D}$ steps of decoding: $\widehat{\mathbb{X}}^{[k]}_{\rm out} = [\widehat{\mathbf{X}}^{kT_{\rm D}+T_{\rm E}},...,\widehat{\mathbf{X}}^{(k+1)T_{\rm D}+T_{\rm E}-1}]\in \R^{N\times T_{\rm D} \times 2}$. 

\textbf{Multiple sampling training strategy.} Although modeling trajectories' uncertainty through the bivariate Gaussian distribution provides more diversity for model prediction, it may bring a problem that inaccurate sampled trajectories will deteriorate the training of the encoding-decoding process by introducing inaccurate next inputs. To provide the model with more stable training, here we use a multiple sampling strategy that samples future trajectories multiple times and selects the most accurate one for subsequent prediction in the training phase. We execute the decoding process $K_{\rm ms}$ times and obtain $K_{\rm ms}$ predicted trajectories. We select the trajectories with minimum ADE or FDE with ground-truth trajectories as the output of the decoder and the input of the next encoding-decoding process in the training.

\subsection{Prediction refinement}
\begin{figure}[t] 
\centering
\includegraphics[width=0.45\textwidth]{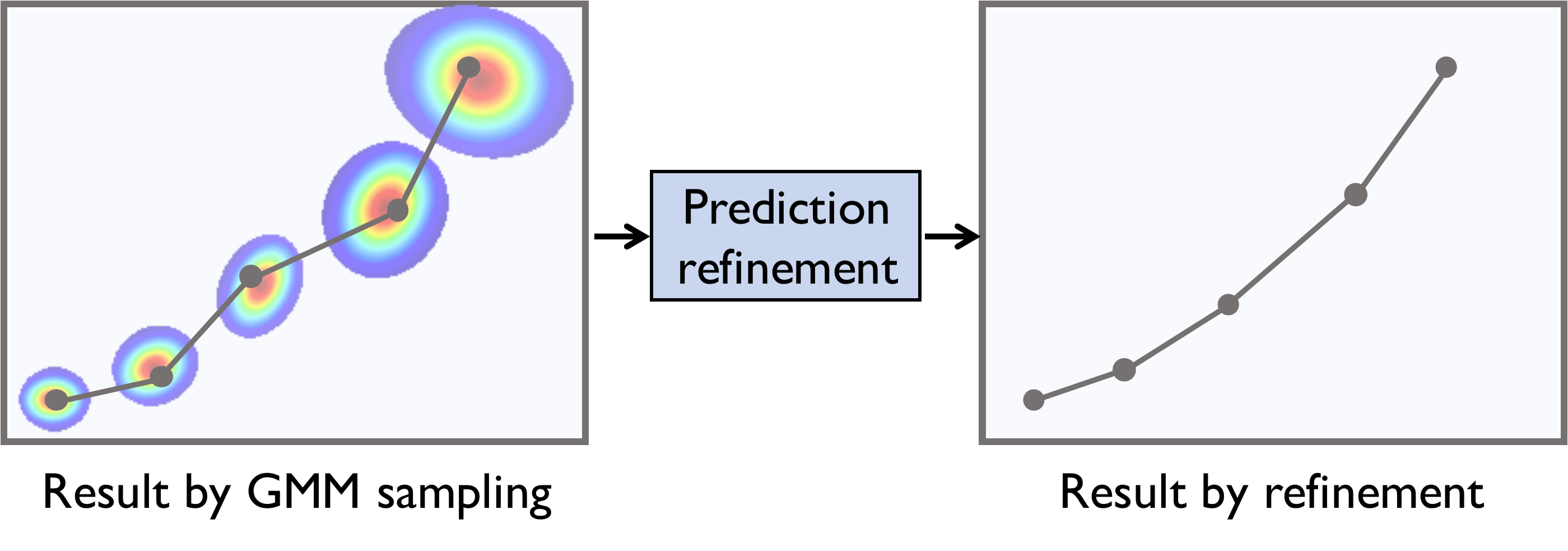}
\vspace{-2mm}
\caption{\small The trajectory obtained by GMM may have the unsmoothing problem since the sampling of every timestamp is independent. Thus we use a predict refinement module to obtain a more smooth and feasible prediction.}
\label{fig:refine}
\vspace{-2mm}
\end{figure}

The encoding-decoding process will repeat $\lceil \frac{T_{\rm f}}{T_{\rm D}}\rceil$ times and we gather all the decoder outputs to obtain the prediction of all future trajectories $\widehat{\mathbb{X}}^{+} = [\widehat{\mathbf{X}}^{T_{\rm p}},\widehat{\mathbf{X}}^{T_{\rm p}+1},\cdots,\widehat{\mathbf{X}}^{T_{\rm p}+T_{\rm f}-1}]$. However, since in the decoding process the GMM model samples the velocity of every timestamps independently, the predicted coordinates may lack of correlation thus lead to problems of unsmoothing, see Fig. \ref{fig:refine} for an example. To enhance the correlation of predicted coordinates across all timestamps, we propose a prediction refinement module to refine the whole future trajectories. Specifically, the prediction refinement can be formulated as $\widehat{\mathbb{X}}^{+}_{\rm ref} = \mathcal{M}_{\rm ref}(\widehat{\mathbb{X}}^{+})$, where $\mathcal{M}_{\rm ref}(\cdot)$ is the refinement function like MLPs. 

\subsection{Training loss}
To train the prediction system, we minimize an overall loss function
$\mathcal{L} = \mathcal{L}_{\mathrm{elbo}} +  \mathcal{L}_{\mathrm{refine}}$, where 
\begin{equation*}
    \begin{aligned}
    &\mathcal{L}_{\mathrm{elbo}} =   
     \beta_1 \sum_{k=0}^{\lceil \frac{T_{\rm f}}{T_{\rm D}}\rceil-1} -\mathbb{E}_{ q(\mathbf{Z}^{[k]}|\widehat{\mathbb{X}}^{[k]}_{\rm in},\mathbb{X}^{[k]}_{\rm out})} \left[\operatorname{log}\; p(\widehat{\mathbb{X}}^{[k]}_{\rm out}|\widehat{\mathbb{X}}^{[k]}_{\rm in},\mathbf{Z}^{[k]})\right] \\
     &\quad\quad\quad +\beta_2 \sum_{k=0}^{\lceil \frac{T_{\rm f}}{T_{\rm D}}\rceil-1} \operatorname{KL}\left(q(\mathbf{Z}^{[k]}|\widehat{\mathbb{X}}^{[k]}_{\rm in},
    \mathbb{X}^{[k]}_{\rm out}) \| p(\mathbf{Z}^{[k]}|\widehat{\mathbb{X}}^{[k]}_{\rm in})\right)
    \end{aligned}
\end{equation*}
is the negative evidence lower-bound with $\beta_1$, $\beta_2$ the hyperparameters and $\operatorname{KL}(\cdot||\cdot)$ the KL divergence, and
\begin{equation*}
    \begin{aligned}
     & \mathcal{L}_{\mathrm{refine}} =   \beta_3 \|\widehat{\mathbb{X}}_{\mathrm{ref}}^{+}-\mathbb{X}^{+}\|_{2}^{2}
    \end{aligned}
\end{equation*}
is the prediction refinement loss with $\beta_3$ the hyperparameter. In the negative evidence lower-bound $\mathcal{L}_{\mathrm{elbo}}$, we add all the negative log-likelihood as well as KL-divergence of every encoding-decoding process. 

\section{Experiments}
\label{sec:experiment}
\subsection{Experimental Setup}
\subsubsection{Datasets}
We evaluate our method on both physical simulation datasets and four real-world public datasets. For the physical simulation datasets, since the ground truth of the interaction relationship can be accessed, we evaluate the ability of relational reasoning of our~\texttt{DynGroupNet}. The generation detail of physical simulation datasets is shown in the appendix. For the real-world datasets, we evaluate the effectiveness of our proposed~\texttt{DynGroupNet}, including:

\textbf{Dataset 1: NBA SportVU Dataset (NBA)} The NBA SportVU dataset\footnote{Data: https://github.com/linouk23/NBA-Player-Movements} contains player and ball trajectories from the 2015-2016 NBA season collected with the SportVU tracking system. The raw tracking data is in the JSON format, and each moment includes information about the identities of the players on the court, the identities of the teams, the period, the game clock, and the shot clock. We selected 50k samples in total for training, validation and testing with a split of 65\%, 10\%, 25\%. Each sample contains the historical 10 timestamps (2.0s) and future 20 timestamps (4.0s).

\textbf{Dataset 2: NFL Football Dataset (NFL)} The NFL Football Dataset\footnote{Data: https://github.com/nfl-football-ops/Big-Data-Bowl} uses NFL’s Next Gen Stats data, which includes identities of the players on the court, the identities of the teams, the time and the position of every player on the field during each play in 2017 year. We both predict the 22 players' (11 player per team) and the ball's trajectories. We selected 13k samples in total for training, validation and testing with a split of 10k, 1k, 2k. Each sample contains the historical 8 timestamps (1.6s) and future 16 timestamps (3.2s).

\textbf{Dataset 3: Stanford Drone Dataset (SDD)} The Stanford Drone dataset (SDD) \cite{robicquet2016learning} consists of 20 scenes captured using a drone in top-down view around the university campus containing several moving agents like humans and vehicles. The coordinates of multiple agents’ trajectories are provided in pixels. we use 0.4s as the time interval, and use the first 3.2 seconds (8 timestamps) to predict the following 4.8 seconds (12 timestamps). We use the standard test train split as used in previous works \cite{sadeghian2019sophie,gupta2018social,mangalam2020not}.

\textbf{Dataset 4: ETH-UCY dataset}
ETH-UCY dataset \cite{lerner2007crowds,pellegrini2009you} contains 5 subsets, ETH, HOTEL, UNIV, ZARA1 and ZARA2. They consist of pedestrian trajectories captured at 2.5Hz in multi-agent social scenarios. Following the experimental setting in \cite{gupta2018social,salzmann2020trajectron++}, we split the trajectories into segments of 8s, where we use 0.4s as the time interval, and use the first 3.2 seconds (8 timestamps) to predict the following 4.8 seconds (12 timestamps). We use the leave-one-out approach, training on 4 sets and testing on the remaining set.

\subsubsection{Metrics}
Following previous works~\cite{li2020evolvegraph,alahi2016social,lee2017desire}, we use $\mathrm{minADE}_K$ and $\mathrm{minFDE}_K$ as evaluation metrics. $\mathrm{minADE}_K$ is the minimum among $K$ average distances of the $K$ predicted trajectories to the ground-truths in terms of the whole trajectories; $\mathrm{minFDE}_K$ shows the minimum distance among $K$ predicted endpoints to the ground-truth endpoints.

\begin{figure*}[t] 
\centering
\includegraphics[width=0.97\textwidth]{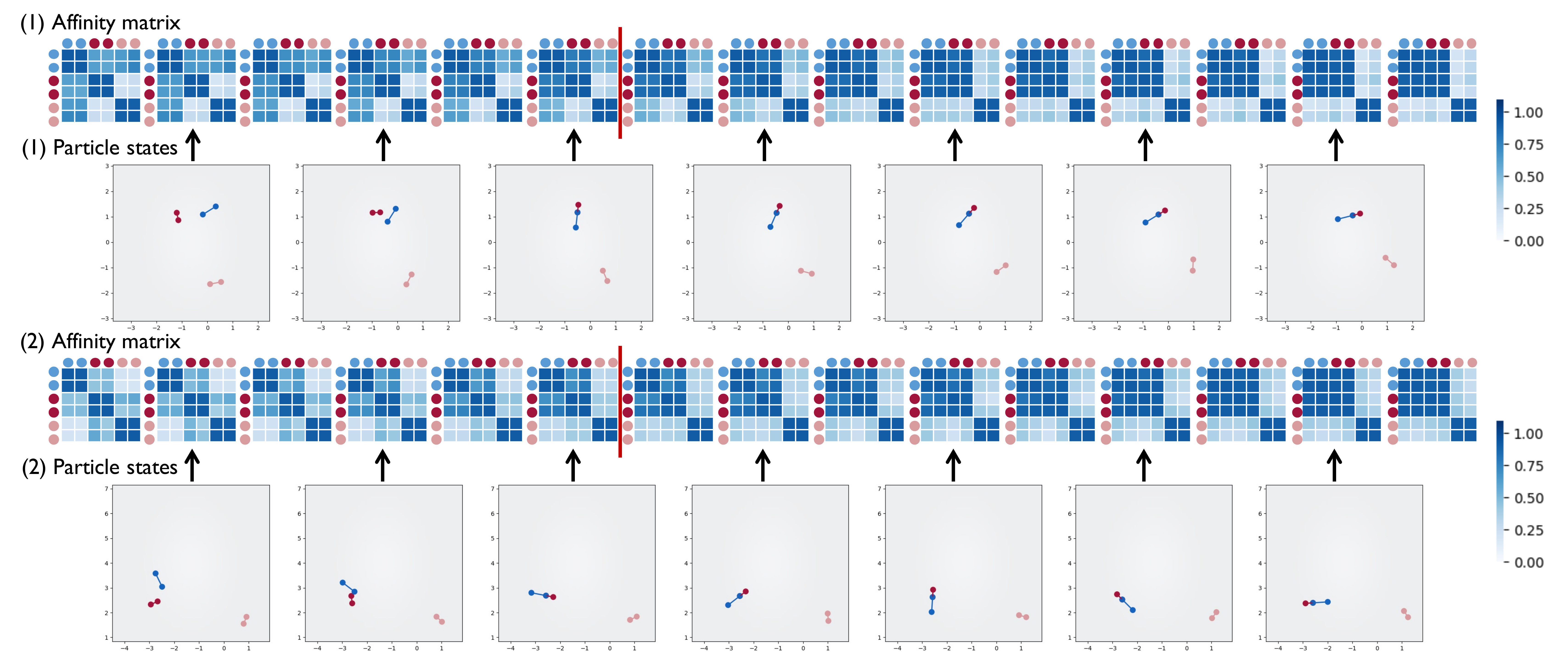}
\vspace{-2mm}
\caption{\small Visualization of learnt dynamic group behavior. 
We visualize the changing of particle connecting states and corresponding learnt affinity matrix of future 15 timestamps. In figures of particles states, ball with the same color are linked by a light bar. At one moment, two particles will collide and merge as one. In figures of affinity matrix heatmaps, a darker color represents a higher affinity score and the red line represents the timestamp when two particles collide.
% There are three 2x2 submatrix before the collision and one 4x4 submatrix and one 2x2 submatrix after the collision in our learnt affinity matrix, reflecting our model captures the group change. 
}
\label{fig:heatmap}
\vspace{-2mm}
\end{figure*}

\subsubsection{Implementation details}
The models are implemented with PyTorch 1.8.0. All the MLPs used the ReLU activation function. For the $0$th scale, we set the $K^{(0)}$ to $N-1$ to obtain a fully connected graph. The latent code dimension $d_z$ is 20 and the dimension of the feature at a single scale $d$ is 32. $\alpha$ in affinity modeling is 0.2, and temperature $\gamma$ in Gumbel softmax is 0.5. The category number $L$ is set to 6 at $0$th scale and 10 at other scales. The multiple-sampling times $K_{ms}$ is set to 5. Each per-category function is implemented as a three-layer MLP with a hidden size of 128. In the dynamic embedding evolvement, we apply a two-layer transformer with 8 heads. The future encoding module $\mathcal{F}_{\rm f}$ is implemented with a 1-layer bidirectional LSTM with a hidden dimension of 32. The GRU used in the decoder has one layer with a hidden dimension of 128. The prediction refinement function is implemented as a two-layer MLP with the hidden dimension of 256 and 512. For NBA dataset, we set the decoding horizon $T_{\rm D}=\tau=5$, scale to 2,5,11 players. For NFL dataset, we set decoding horizon $T_{\rm D}=\tau=8$, the scale to 2,3,4,5,6,23 players. For the SDD dataset, we set the decoding horizon $T_{\rm D}=\tau=4$, group scales to 2 and full of agent number in less than 100 pixels distance and. For the ETH dataset, we set decoding horizon $T_{\rm D}=\tau=4$, the group scales to 2 and the full of agents number in less than 5 meters. To train the model, we apply Adam optimizer \cite{kingma2014adam} with an initial learning rate $10^{-4}$ and decay every 10 epochs for 150 epochs on one GTX-3090Ti GPU. For a stable training of refinement model, we set the weight $\beta_1=\beta_2=1.0,\beta_3=0$ in the first 80 epochs and $\beta_1=\beta_2=\beta_3=1.0$ in the last 70 epochs in the loss function.

\subsection{Validation on relational reasoning}
\subsubsection{Ability to capture dynamic group behaviors.}
Here we consider six particles two of each are connected by a light bar first, forming three 2-groups. The light bar will rotate and translate by random angular velocity and translation velocity. At one future moment, the connection relationship will change. Two particles connected by different light bars will collide together and the two particles merge as one. Thus the two light bar will splice together connecting 4 particles to form a 4-group.
The collision process follows the conservation of angular momentum principle, see particle states in Fig. \ref{fig:heatmap} for two examples. We predict the particle states at future 15 timestamps based on the observations of 10 timestamps. We visualize our learnt corresponding affinity matrix across time in Fig. \ref{fig:heatmap}. The red line indicates the moment when two particles collide together. We see that i) at first, the affinity matrix mainly forms three $2\time2$ sub-matrices representing three 2-groups; and ii) when two particles collide together, affinity scores between red and blue particles increase and the affinity matrix turns to forming one $4\times4$ sub-matrix and one $2\time2$ sub-matrix, representing one 4-group and a 2-group, indicating that we effectively capture the dynamic group behavior across time.

\begin{figure*}[t] 
\centering
\includegraphics[width=0.9\textwidth]{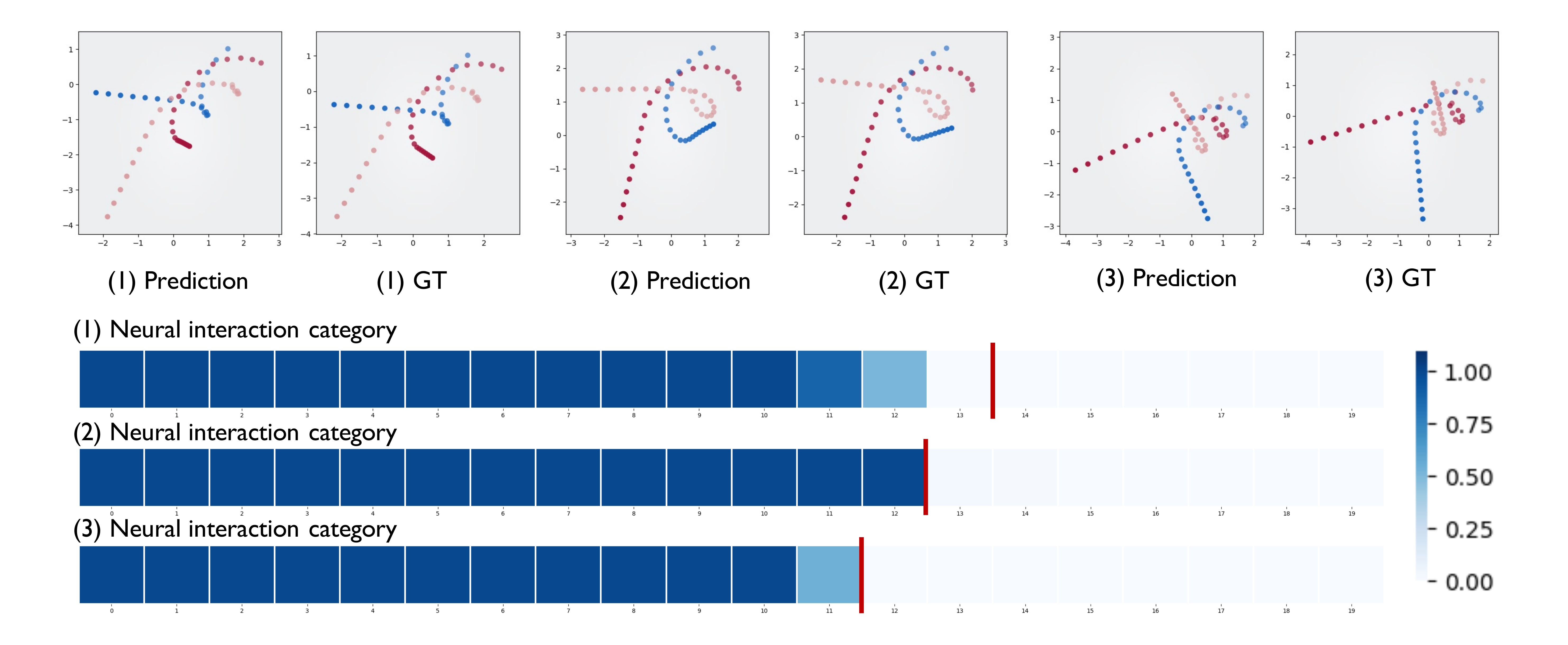}
\vspace{-2mm}
\caption{\small Visualization of neural interaction category. Top: particle trajectories of predicted and ground-truth trajectories of three data samples. Three balls are connected by a Y-shape light bar at first and disconnected at one specific timestamp. Bottom: Heatmaps of our reasoned neural interaction category of 20 future timestamps. A darker color represents a higher probability of particles being linked and the red line represents the timestamp when particles are disconnected.}
\label{fig:category}
\vspace{-2mm}
\end{figure*}

\begin{figure*}[t] 
\centering
\includegraphics[width=0.9\textwidth]{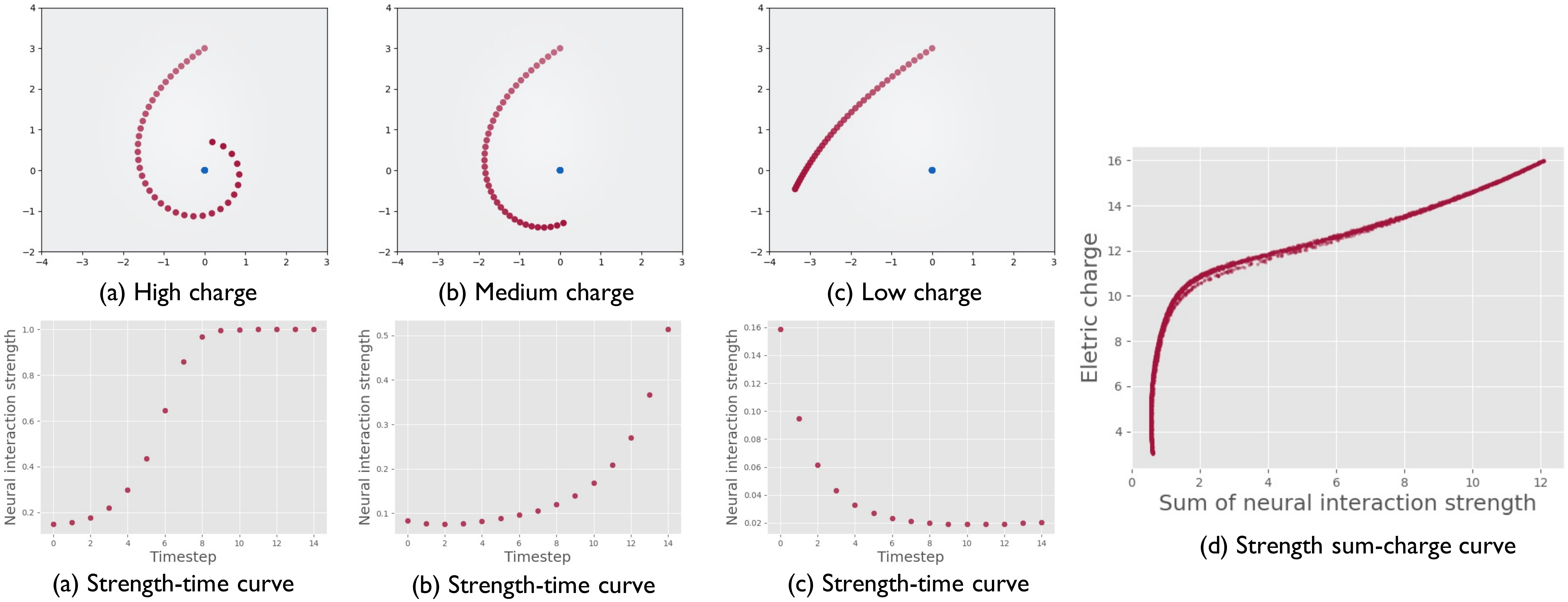}
\vspace{-2mm}
\caption{\small Visualization of neural interaction strength. The red positively charged particle will receive the attractive force from the blue negatively charged particle. (a)(b)(c) present three data samples under situations that the blue particle carries high charge, medium charge and low charge, and their corresponding relation curves of learnt neural interaction strength with timestamps. (d) presents the relation curve of the sum of learnt neural interaction strengths and the amount of charge that the blue particle carries.}
\label{fig:strength}
\end{figure*}

\subsubsection{Ability to reason dynamic interaction category}
Here we consider three particles forming one group with two possible interaction categories: connected and disconnected. At first, three particles are connected by a Y-shape light bar to have a rotation and translation movement. At the moment when the center of the three particles reaches cross the line $x=0$, the light bar disappears and particles move freely with their velocity at the disappearing moment. We predict the particle states at future 20 timestamps based on the observations of 10 timestamps. Fig. \ref{fig:category} gives three examples.

We compare the neural interaction category in ~\eqref{eq:decouple} with the ground-truth category at every timestamp to achieve dynamic interaction recognition and report the recognition accuracy. This category recognition is unsupervised because the model only accesses the trajectories of particles without any prior information about the three types of interaction. We compare our method with three baselines in NRI~\cite{kipf2018neural} and dNRI~\cite{graber2020dynamic} which consider only pair-wise interactions and an upper bound. To adapt them to group-wise actions, we use a majority vote for all pair-wise edges. The upper bound `Supervised' represents using the ground-truth category to train the neural interaction category. TABLE~\ref{table:type} reports the recognition accuracy on static (the interaction category keeps the same in all timestamps) and dynamic interaction recognition tasks, which are averaged over 5 independent runs. `--' represents not applicable. We see that i) our method significantly outperforms four baselines, as we model the group interaction while NRI only considers pair-wise interaction; and ii) our method achieves a close performance to the upper bound. We also visualize our learnt time-varying neural interaction category in Fig. \ref{fig:category}. The red line represents the moment when the light bar disappears that bringing the interaction category change. We see that our method can quickly capture the change of the interaction category within one timestamp. Both the numerical results and the visualization results reflect that our method is capable of reasoning the interaction category in an unsupervised manner.

\begin{table}[!t]
    \centering
    \caption{\small Comparison of accuracy (mean $\pm$ std in \%) of interaction category recognition.}
    \vspace{-2mm}
    \small
    \setlength{\tabcolsep}{0.4mm}{
    \begin{tabular}{c|ccccc|c}
        \hline
        \hline
        ~ &\makecell{Corr.\\(path)} &\makecell{Corr.\\(LSTM)} & NRI   & dNRI & \textbf{Ours}  & Supervised \\
        \hline
        Static &78.3{\scriptsize $\pm$0.0}&75.3{\scriptsize$\pm$2.5} & 89.7{\scriptsize$\pm$3.5}&86.4{\scriptsize$\pm$2.8} & \textbf{99.3}{\scriptsize$\pm$0.3}&99.5{\scriptsize$\pm$0.2} \\
        Dynamic &\textbf{--}&\textbf{--} & 72.1{\scriptsize$\pm$2.4} & 77.2{\scriptsize$\pm$2.3}& \textbf{86.3}{\scriptsize$\pm$2.1}&93.5{\scriptsize$\pm$0.7} \\
        \hline
        \hline
    \end{tabular}}
    \label{table:type}
\vspace{-3mm}
\end{table}

\begin{table*}[t]
\centering
\setlength{\tabcolsep}{1mm}\caption{minADE$_{20}$/minFDE$_{20}$ (meters) of trajectory prediction (NBA dataset). $*$ denotes that is the previous best model on this dataset and the \textbf{bold} font represents the best result. Our~\texttt{DynGroupNet} achieves a $22.6\%/28.0\%$ ADE/FDE improvement compared to Trajectron++.}
\vspace{-2mm}
% \fontsize{8}{8}\selectfont
\setlength{\tabcolsep}{2mm}{
\resizebox{\textwidth}{!}{
\begin{tabular}{l|ccccccccc|cc}
    \hline
    \hline
    Time
    & \makecell[c]{Social\\-LSTM\cite{alahi2016social}}
    % &\makecell[c]{Social-\\Attention\cite{vemula2018social}}
    &\makecell[c]{Social\\-GAN\cite{gupta2018social}} &\makecell[c]{Social-\\STGCNN\cite{mohamed2020social}} 
    &\makecell[c]{STGAT\\\cite{huang2019stgat}}
    & \makecell[c]{NRI\\\cite{kipf2018neural}}
    & \makecell[c]{STAR\\\cite{yu2020spatio}}
    % & \makecell[c]{Trajectron++\\\cite{salzmann2020trajectron++}}
    % \makecell[c]{EvolveGraph\\\cite{li2020evolvegraph}} &
    & 
    \makecell[c]{PECNet\\\cite{mangalam2020not}} &
    \makecell[c]{NMMP\\\cite{hu2020collaborative}}
    &
    \makecell[c]{Trajectron++*\\\cite{salzmann2020trajectron++}}
    &\makecell[c]{DynGroupNet \\ (static)}  &\makecell[c]{DynGroupNet} 
    \\
\hline
     1.0s &0.42/0.63& 0.41/0.62&   0.34/0.48 &0.35/0.51 & 0.43/0.61 &0.43/0.66&0.45/0.79 &0.35/0.51 &0.30/0.38&0.23/0.33 &\textbf{0.19}/\textbf{0.28}  \\
     2.0s &0.79/1.47 &0.81/1.32& 0.71/0.94  & 0.73/1.10& 0.81/1.35& 0.75/1.24& 0.96/1.66&0.65/1.07&0.59/0.82&0.47/0.71&\textbf{0.40}/\textbf{0.61} \\
     3.0s & 1.21/2.15 &1.19/1.94&1.09/1.77&1.04/1.75&1.14/2.15 & 1.03/1.51& 1.46/2.54&0.98/1.59&0.85/1.24&0.71/1.03 &\textbf{0.65}/\textbf{0.90}  \\
     Total(4.0s) & 1.65/2.98&1.59/2.41& 1.53/2.26 & 1.40/2.18&1.59/2.73&1.13/2.01 &1.86/3.47&1.31/2.03 &1.15/1.57 &0.95/1.31 &\textbf{0.89}/\textbf{1.13}                     \\
    \hline
    \hline
\end{tabular}}
}
\label{table:nba}
\vspace{-2mm}
\end{table*}

\subsubsection{Ability to reason dynamic interaction strength.}
We consider the movement of two charged particles that interact via Coulomb forces. One fixed center particle brings a random negative charge and another moving particle brings a fixed positive charge thus there exists an attractive force between the two particles. The positive particle has an initial velocity and will encounter air resistance proportional to its speed. When the center particle brings a high charge, the moving particle will be attracted to the center particle and when the center particle brings a low charge, the moving particle will escape since the attractive force is small. We predict the particle states of 15 future timestamps based on the observations of 25 timestamps. Fig.~\ref{fig:strength}(a)(b)(c) shows examples of particle trajectories in the case of high charge, medium charge and low charge. A stronger attraction the moving particle receives leads to a higher interaction strength. During the prediction, we aim to validate whether the dynamic neural interaction strength in~\eqref{eq:decouple} can reflect the time-varying attraction force and the amount of charge without any direct supervision.

Fig.~\ref{fig:strength}(a)(b)(c) shows the relations between timestamps ($x$-axis) and the learnt dynamic interaction strength ($y$-axis) in three cases: high, medium and low. We also show the relations between the sum of learnt dynamic interaction strength of data samples ($x$-axis) and their corresponding charge on the fixed particle ($y$-axis) in Fig.~\ref{fig:strength}(d), using 1000 data samples. 
We see that i) in the cases of high and medium charge, the moving particle is attracted by the center particle. The distance between two particles is decreasing and thus the moving particle receives higher and higher Coulomb attraction forces, leading to an increasing interaction strength. In these two cases, our learnt neural interaction strength also increases over time; ii) in the case of low charge, the moving particle escapes. The distance between two particles is increasing and thus the moving particle receives lower and lower Coulomb attraction forces, leading to a decreasing interaction strength. In this case, our learnt neural interaction strength also decreases over time; and iii) the sum of learnt neural interaction strength has a proportional relationship with the amount of charge. All the three results reflect our model is capable of capturing the dynamic interaction strength across time in an unsupervised manner.

\begin{table*}[t]
\centering
\setlength{\tabcolsep}{1mm}\caption{minADE$_{20}$/minFDE$_{20}$ (meters) of trajectory prediction (NFL Football dataset). $*$ denotes that is the previous best model on this dataset and  the \textbf{bold} font represents the best result. Our~\texttt{DynGroupNet} achieves a $26.9\%/34.9\%$ ADE/FDE improvement compared to NMMP.}
\vspace{-2mm}
% \fontsize{8}{8}\selectfont
\setlength{\tabcolsep}{2mm}{
\resizebox{\textwidth}{!}{
\begin{tabular}{l|cccccccccc|cc}
    \hline
    \hline
    Time
    % &\makecell[c]{Social-\\Attention\cite{vemula2018social}}
    &\makecell[c]{Social\\-GAN\cite{gupta2018social}} &\makecell[c]{Social-\\STGCNN\cite{mohamed2020social}} 
    &\makecell[c]{STGAT\\\cite{huang2019stgat}}
    & \makecell[c]{NRI\\\cite{kipf2018neural}}
    & \makecell[c]{STAR\\\cite{yu2020spatio}}
    % & \makecell[c]{Trajectron++\\\cite{salzmann2020trajectron++}}
    % \makecell[c]{EvolveGraph\\\cite{li2020evolvegraph}} &
    & 
    \makecell[c]{PECNet\\\cite{mangalam2020not}} 
    &\makecell[c]{LB-EBM\\\cite{pang2021trajectory}}
    &\makecell[c]{DNRI\\\cite{graber2020dynamic}}&
    \makecell[c]{Trajectron++\\\cite{salzmann2020trajectron++}}
    &
    \makecell[c]{NMMP*\\\cite{hu2020collaborative}}
    &\makecell[c]{DynGroupNet \\ (static)}  &\makecell[c]{DynGroupNet} 
    \\
\hline
     1.0s &0.37/0.68&0.45/0.64&0.35/0.64&0.63/0.86&0.49/0.84&0.52/0.97&0.75/1.05& 0.49/0.92 & 0.41/0.65 &0.34/0.61& 0.33/0.45&\textbf{0.25}/\textbf{0.37} \\
     2.0s &0.83/1.53 &1.06/1.87&0.82/1.60&1.17/2.14&1.02/1.84&1.19/2.47&1.26/2.28& 1.26/2.57 & 0.93/1.65&0.78/1.48 &0.66/0.99&\textbf{0.56}/\textbf{0.94} \\
     Total(3.2s) & 1.44/2.51 &1.82/3.18&1.39/2.48&1.90/3.44&1.51/2.97&1.99/3.84&1.90/3.25& 2.25/4.37 & 1.54/2.58&1.30/2.29 &1.06/1.56&\textbf{0.95}/\textbf{1.49}  \\
    \hline
    \hline
\end{tabular}}
}
\label{table:nfl}
\vspace{-1mm}
\end{table*}

\begin{table*}[!t]
\footnotesize
\centering
\setlength{\tabcolsep}{1mm}{\caption{minADE$_{20}$/minFDE$_{20}$ (pixels) of trajectory prediction (SDD dataset). $*$ denotes that is the previous best model on this dataset and the \textbf{bold} font represents the best result. Our~\texttt{DynGroupNet} achieves a $5.1\%/13.0\%$ ADE/FDE improvement compared to LB-EBM.}
\vspace{-2mm}
% \fontsize{8}{8}\selectfont
\resizebox{\textwidth}{!}{
\begin{tabular}{l|ccccccccccc|cc}
\hline
\hline
    Time 
    &\makecell[c]{Social-\\LSTM\cite{alahi2016social}}
    & \makecell[c]{Social\\-GAN\cite{gupta2018social}}
    % &\makecell[c]{Social-\\STGCNN\cite{mohamed2020social}}
    &\makecell[c]{SOPHIE\\\cite{sadeghian2019sophie}}  & \makecell[c]{Trajectron++\\\cite{salzmann2020trajectron++}}
    & \makecell[c]{NMMP\cite{hu2020collaborative}}
    & \makecell[c]{EvolveGraph\\\cite{li2020evolvegraph}}&CF-VAE \cite{bhattacharyya2019conditional}
    &\makecell[c]{MG-GAN\\\cite{dendorfer2021mg}}
    &\makecell[c]{AgentFormer\\\cite{yuan2021agentformer}}
    &\makecell[c]{PECNet\\\cite{mangalam2020not}}   &
    \makecell[c]{LB-EBM*\\\cite{pang2021trajectory}} &\makecell[c]{DynGroupNet \\ (static)}  &\makecell[c]{DynGroupNet} 
    \\
    % &\makecell[c]{\textbf{Ours}\\(Attention)} \\
\hline
     4.8s& 31.19/56.97 & 27.23/41.44
     & 16.27/29.38
     &19.30/32.70&14.67/26.72&13.90/22.90&12.60/22.30&13.60/25.80&10.32/18.30&9.96/15.88&8.87/15.61&9.58/16.19 &\textbf{8.42}/\textbf{13.58} \\
    %  &\textbf{9.72}/\textbf{15.45} \\
\hline
\hline
\end{tabular}
\label{table:sdd}}}
\vspace{-1mm}
\end{table*}

\begin{table*}[!t]
\centering
\setlength{\tabcolsep}{1mm}\caption{minADE$_{20}$/minFDE$_{20}$ (meters) of trajectory prediction (ETH-UCY dataset).  $*$ denotes that is the previous best model on this dataset and the \textbf{bold}/\underline{underlined} font represents the best/second best result. In every subset our~\texttt{DynGroupNet} achieves the best or close to the best performance in ADE/FDE and our method achieves comparable averaged results compared to LB-EBM.}
% \fontsize{8}{8}\selectfont
\vspace{-2mm}
\resizebox{\textwidth}{!}{
\begin{tabular}{l|ccccccccccccccc|cc}
  \hline
  \hline
    Subset & \makecell[c]{Social-\\LSTM\cite{alahi2016social}}& \makecell[c]{Group-\\LSTM\cite{bisagno2018group}} &\makecell[c]{Social-\\GAN\cite{gupta2018social}}&\makecell[c]{SOPHIE\\\cite{sadeghian2019sophie}}&\makecell[c]{STGAT \\\cite{huang2019stgat}}&\makecell[c]{NMMP\\\cite{hu2020collaborative}}&\makecell[c]{STAR\\\cite{yu2020spatio}}&\makecell[c]{PECNet\\\cite{mangalam2020not}} &\makecell[c]{Trajectron++\\\cite{salzmann2020trajectron++}}&\makecell[c]{MG-GAN\\\cite{dendorfer2021mg}}&\makecell[c]{Causal-\\STGAT\cite{chen2021human}}&
    \makecell[c]{Grouptron\\\cite{zhou2021grouptron}}&\makecell[c]{GA-STT\\\cite{zhou2022ga}}& \makecell[c]{AgentFormer\\\cite{yuan2021agentformer}}&\makecell[c]{LB-EBM*\\\cite{pang2021trajectory}}& \makecell[c]{DynGroupNet\\(static)}& \makecell[c]{DynGroupNet} \\
    % &\makecell[c]{\textbf{Ours}\\(Attention)} \\
\hline
     ETH& 1.09/2.35&0.28/1.12&0.81/1.52&0.70/1.43&0.65/1.12&0.61/1.08 &0.36/0.65&0.54/0.87 &0.61/1.02&0.47/0.91&0.60/0.98&0.70/1.56&0.51/0.89&0.45/0.75&\textbf{0.30}/\textbf{0.52}&0.44/0.74&\underline{0.42}/\underline{0.66} \\
     HOTEL&0.79/1.76&0.28/0.89&0.72/1.61&0.76/1.67&0.35/0.66&0.33/0.63&0.17/0.36&0.18/0.24&0.19/0.28&\underline{0.14}/0.24&0.30/0.54&0.21/0.46&0.22/0.46&\underline{0.14}/\underline{0.22}&\textbf{0.13}/\textbf{0.20}&\textbf{0.13}/\textbf{0.20}&\textbf{0.13}/\textbf{0.20}\\
     UNIV&0.67/1.40&0.56/1.48&0.60/1.26&0.54/1.24&0.52/1.10&0.52/1.11&0.31/0.62&0.35/0.60&0.30/0.54&0.54/1.07&0.52/1.10&0.38/0.97&0.29/0.63&\underline{0.25}/\underline{0.45}&0.27/0.52&0.26/0.50&\textbf{0.24}/\textbf{0.44}  \\
     ZARA1&0.47/1.00&0.23/0.91&0.34/0.69&0.30/0.63&0.34/0.69&0.32/0.66&0.26/0.55&0.22/0.39 &0.24/0.42&0.36/0.73&0.32/0.64&0.30/0.76&0.25/0.55&\textbf{0.18}/\textbf{0.30}&0.20/0.37&0.22/0.43&\underline{0.19}/\underline{0.34}\\
     ZARA2&0.56/1.17&0.34/1.49&0.42/0.84&0.38/0.78&0.29/0.60&0.43/0.85&0.22/0.46&0.17/0.30&0.18/0.31&0.29/0.60&0.28/0.58&0.22/0.56&0.20/0.44&\textbf{0.14}/\textbf{0.24} &\underline{0.15}/0.29&\underline{0.15}/\underline{0.28}&\underline{0.15}/\underline{0.28}\\
     AVG&  0.72/1.54&0.34/1.18&0.58/1.18&0.54/1.15&0.43/0.83&0.41/0.82&0.26/0.53&0.29/0.48&0.30/0.51&0.36/0.71&0.40/0.77&0.36/0.86&0.29/0.59&\underline{0.23}/0.39&\textbf{0.21}/\textbf{0.38}&0.24/0.43&\underline{0.23}/\textbf{0.38}\\
  \hline
  \hline
\end{tabular}}
\vspace{-1mm}
\label{table:eth}
\end{table*}

\subsection{Validation on trajectory prediction}
\mypar{NBA dataset} We predict the future 20 timestamps (4.0s) based on the historical 10 timestamps (2.0s). TABLE \ref{table:nba} shows the comparison with nine state-of-the-art methods of ADE/FDE metrics at different times. The state-of-the-art method are based on recurrent networks \cite{alahi2016social}, attention mechanism \cite{gupta2018social,yu2020spatio,huang2019stgat,mangalam2020not} and graph-based mechanism \cite{mohamed2020social,kipf2018neural,hu2020collaborative,salzmann2020trajectron++}. Moreover, to investigate the effect of the dynamic multiscale hypergraph, we also report the result that uses a static multiscale hypergraph DynGroupNet (static) by setting the decoding length to the length of future trajectories $T_{\rm D} = T_{\rm f} = 20$. We see that i)~\texttt{DynGroupNet} with our proposed framework achieves a significant improvement over the state-of-the-art methods: the ADE/FDE at 4.0s is reduced by $22.6\%/28.0 \%$ compared to the best baseline method (Trajectron++). The improvement over previous methods increases with timestamps since the proposed dynamic multiscale hypergraph can capture more comprehensive interaction patterns across time; and ii) the performance gains when applying a dynamic multiscale hypergraph comparing to a static multiscale hypergraph because the dynamic multiscale hypergraph is capable to capture the group belongingness and interaction change across time, which is common in basketball games with rapid tactical changes.

\mypar{NFL Football dataset}
We predict the future 16 timestamps (3.2s) based on the historical 8 timestamps (1.6s). TABLE \ref{table:nfl} shows the comparison with ten state-of-the-art methods of ADE/FDE metrics at different times. We also report the result that uses a static multiscale hypergraph DynGroupNet (static) by setting the decoding length to the length of future trajectories $T_{\rm D} = T_{\rm f} = 12$. We see that i)~\texttt{DynGroupNet} with our proposed framework achieves a significant improvement over the state-of-the-art methods: the ADE/FDE at 3.2s is reduced by $26.9\%/34.9\%$ comparing to the best baseline method (NMMP); and ii) the performance gains when applying a dynamic multiscale hypergraph comparing to a static multiscale hypergraph, proving the effectiveness of our dynamic multiscale hypergraph.

\mypar{SDD dataset}
We predict the future 12 timestamps (4.8s) based on the historical 8 timestamps (3.2s). TABLE~\ref{table:sdd} compares the proposed method with eleven state-of-the-art methods. We also report the result that uses a static multiscale hypergraph DynGroupNet (static) by setting the decoding length to the length of future trajectories $T_{\rm D} = T_{\rm f} = 12$. We see that i) our~\texttt{DynGroupNet} achieves state-of-the-art performance and reduces the ADE/FDE by $5.1\%/13.0\%$ comparing to the previous best method (LB-EBM); and ii) using a dynamic multiscale hypergraph achieve a better performance comparing to static multiscale hypergraph, proving the effectiveness of our dynamic multiscale hypergraph.

\mypar{ETH-UCY dataset}
We predict the future 12 timestamps (4.8s) based on the historical 8 timestamps (3.2s). TABLE~\ref{table:eth} compares the proposed method with fifteen state-of-the-art methods and the `AVG' means the averaged result over 5 subsets. We also report the result that uses a static multiscale hypergraph DynGroupNet (static) by setting the decoding length to the length of future trajectories $T_{\rm D} = T_{\rm f} = 12$.
We see that i)~\texttt{DynGroupNet} outperforms most of the previous methods and achieves comparable average ADE/FDE performance with the state-of-the-art method (LB-EBM) and achieves the best FDE performance; ii) in every subset, our method achieves the best or close to the best performance in ADE/FDE; iii)~\texttt{DynGroupNet} significantly outperforms all the existing methods that consider group interactions including Group-LSTM, Grouptron, and GA-STT; and iv) using a dynamic multiscale hypergraph achieves a better performance comparing to a static multiscale hypergraph on most of subsets, proving the effectiveness of our dynamic multiscale hypergraph. 

\begin{table}[t]
    \caption{\small Ablation studies of different group scales on the NBA dataset. The scale means number of agents in the groups. We report minADE$_{20}$/minFDE$_{20}$ (meters).}
    \vspace{-2mm}
    \centering
    \small
    \setlength{\tabcolsep}{1mm}{
   \begin{tabular}{c|cccc}
    \hline
    \hline
         & \multicolumn{4}{c}{Prediction time} \\
    \hline
        Scale&1.0s&2.0s&3.0s&4.0s \\
    \hline

        1& 0.23/0.34 & 0.48/0.77 & 0.76/1.14 & 1.02/1.42\\
        1,2 & 0.20/0.29&0.44/0.66&0.68/0.97&0.93/1.22 \\
        1,2,5& 0.20/0.28 & 0.42/0.63 &0.65/0.93 &0.89/1.15 \\
        1,2,5,11& \textbf{0.19}/\textbf{0.28}&\textbf{0.40}/\textbf{0.61}&\textbf{0.65}/\textbf{0.90}&\textbf{0.89}/\textbf{1.13}\\
        1,2,3,5,11&0.19/0.28&0.40/0.62&0.65/0.90&0.90/1.13 \\
    \hline
    \hline
    \end{tabular}
    \label{table:scales_nba}
  }
\end{table}

\begin{table}[t]
    \caption{\small Ablation studies of different group scales on the NFL Football dataset. The scale means number of agents in the groups. We report minADE$_{20}$/minFDE$_{20}$ (meters). }
    \vspace{-2mm}
    \centering
    \small
    \setlength{\tabcolsep}{3mm}{
   \begin{tabular}{c|ccc}
    \hline
    \hline
         & \multicolumn{3}{c}{Prediction time} \\
    \hline
        Scale&1.0s&2.0s&3.2s(Total) \\
    \hline
        1& 0.27/0.42& 0.63/1.09 & 1.10/1.84 \\
        1,2& 0.27/0.39& 0.59/1.00 & 1.00/1.60\\
        1,2,4& 0.26/0.39 & 0.58/0.96 & 0.98/1.54 \\
        1,2,4,5& 0.26/0.38 & 0.57/0.96 & 0.98/1.53\\
        1,2,4,5,23& 0.25/0.38& 0.57/0.96 & 0.96/1.52\\
        1,2,4,5,6,23& 0.25/0.38&  0.56/0.94 & 0.95/1.50\\
        1,2,3,4,5,6,23& \textbf{0.25}/\textbf{0.37}& \textbf{0.56}/\textbf{0.94} & \textbf{0.95}/\textbf{1.49}\\
        1,2,4,5,6,11,23& 0.25/0.37& 0.56/0.94 & 0.96/1.50\\
        % 1,2,4,5&\\
        % 1,2,4,5,23&\\
        % 1,2,23& 0.26/0.38 & 0.57/0.94 & 0.96/1.51\\
        % 1,2,5,23& 0.25/0.38 & 0.57/0.96 & 0.96/1.53\\
        % 1,2,4,5,23& 0.25/0.38& 0.57/0.96 & 0.96/1.53\\
        % 1,2,4,5,6,23& 0.25/0.38&  0.56/0.94 & 0.95/1.50\\
        % 1,2,4,5,6,11,23& 0.26/0.38&  0.56/0.95 & 0.96/1.51\\
        % 1,2,4,5,11,23& 0.54/0.75& 0.96/1.46 & 1.42/1.99\\
        % 1,2,4,5,11,23& 0.54/0.75& 0.96/1.46 & 1.42/1.99\\
        % 1,2,4,5,9,23& 0.26/0.38& 0.57/0.95 & 0.97/1.52\\
        % 1,2,4,5,15,23& 0.26/0.38& 0.57/0.95 & 0.97/1.52\\
        % 1,2,3,4,5,6,23& 0.25/0.37& 0.56/0.94 & 0.95/1.49\\
    \hline
    \hline
    \end{tabular}
    \label{table:scales_nfl}
  }
  \vspace{-2mm}
\end{table}

\begin{figure}[!t]
\centering
\subfigure[\small NBA]{
\begin{minipage}[t]{0.49\linewidth}
\centering
\includegraphics[width=1\textwidth,height=0.73\textwidth]{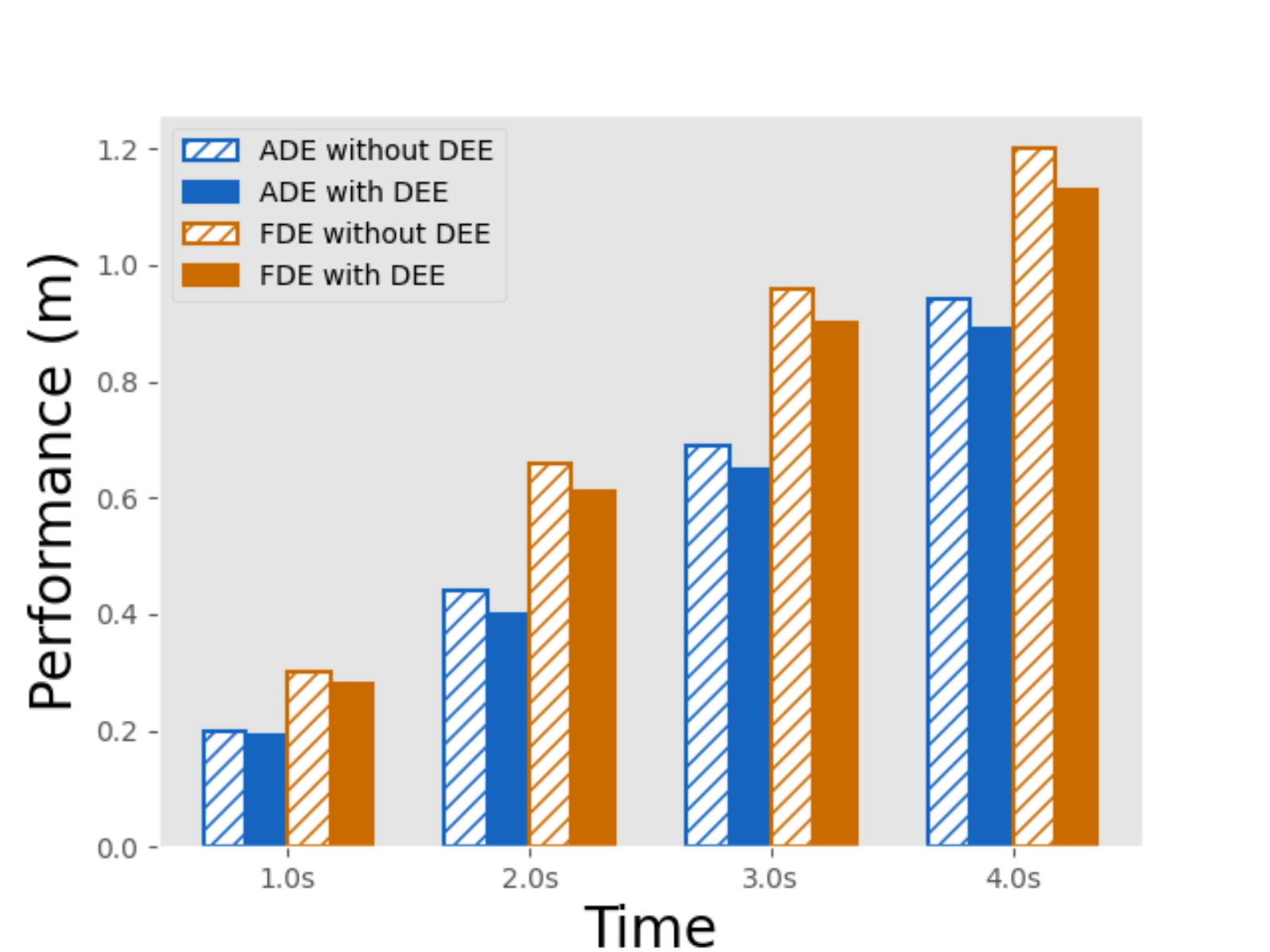}
% \caption{fig1}
\end{minipage}%
}%
\subfigure[\small NFL Football]{
\begin{minipage}[t]{0.49\linewidth}
\centering
\includegraphics[width=1\textwidth,height=0.73\textwidth]{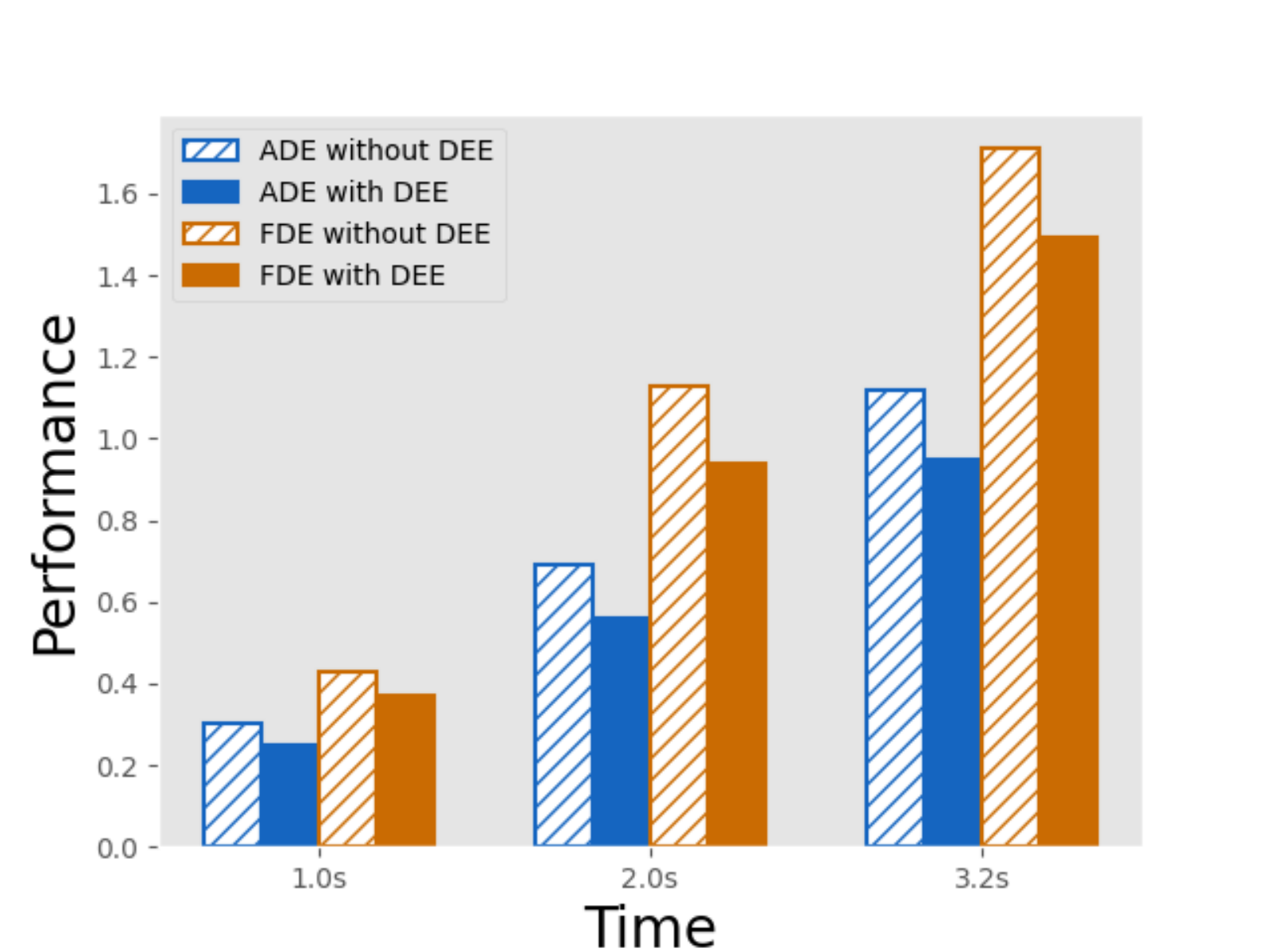}
%\caption{fig2}
\end{minipage}
}%
\centering
\vspace{-3mm}
\caption{minADE$_{20}$ (blue)/minFDE$_{20}$ (orange) of without (shadow bar)/with (solid bar) dynamic embedding evolvement (DEE) on the NBA dataset and the NFL Football dataset.}
\label{fig:ablation_dfe}
\end{figure}

\subsection{Ablation studies}
\subsubsection{Ablation on scales}
TABLE \ref{table:scales_nba} and TABLE \ref{table:scales_nfl} present the effect of multiple scales in the~\texttt{DynGroupNet} on the NBA dataset and the NFL Football dataset. Here the scale means the number of agents in the groups. We see that i) The performance increases at first when choosing more scales; and ii) the performance becomes stable when having sufficient scales, which indicates that there are sufficient group-wise interactions being modeled. 

\begin{table}[!t]
    \caption{\small  Ablation studies of different evolving gap $\tau$ \& decoding length $T_{\rm D}$ on the NBA dataset. We report minADE$_{20}$/minFDE$_{20}$ (meters). }
    \vspace{-2mm}
    \centering
    \setlength{\tabcolsep}{1.5mm}{\small
\begin{tabular}{c|cccc}
\hline
\hline
% \multicolumn{1}{c|}{$\mathcal{M}$} &
\multicolumn{1}{c|}{} & \multicolumn{4}{c}{Prediction time}  \\ \hline
\makecell[c]{$\tau$ \&  $T_{\rm D}$} & 1.0s & 2.0s & 3.0s & 4.0s \\
\hline
1  & 0.20/0.28&	0.42/0.63&	0.66/0.93&	0.90/1.19  \\
% MLP   & ATT   &  0.36/0.51  &  0.65/1.03    &  0.94/2.46    & 1.23/1.91    \\
% MLP  & GroupNet & 0.36/0.51&0.66/1.03&0.95/1.47&1.23/1.91    \\
2 & 0.20/0.28&	0.42/0.62&	0.66/0.93&	0.90/1.16\\
4   &   0.21/0.28&	0.42/0.62&	0.65/0.91&	0.89/1.14\\
% GroupNet  & MLP          &     0.35/0.49    &   0.62/0.95   &   0.88/1.32   & 1.14/1.70   \\
% GroupNet  & ATT &     0.35/0.49    &   0.63/0.95   &   0.88/1.32   & 1.14/1.70   \\
5   &   \textbf{0.19}/\textbf{0.28}&\textbf{0.40}/\textbf{0.61}&\textbf{0.65}/\textbf{0.90}&\textbf{0.89}/\textbf{1.13} \\
10   &   0.21/0.30&0.43/0.66&0.68/0.96&0.92/1.20 \\
20   &   0.23/0.33&0.47/0.71&0.71/1.03&0.95/1.31 \\
         \hline
          \hline
\end{tabular}}
\label{table:timestamps}
\end{table}

\begin{figure*}[t] 
\centering
\includegraphics[width=0.98\textwidth]{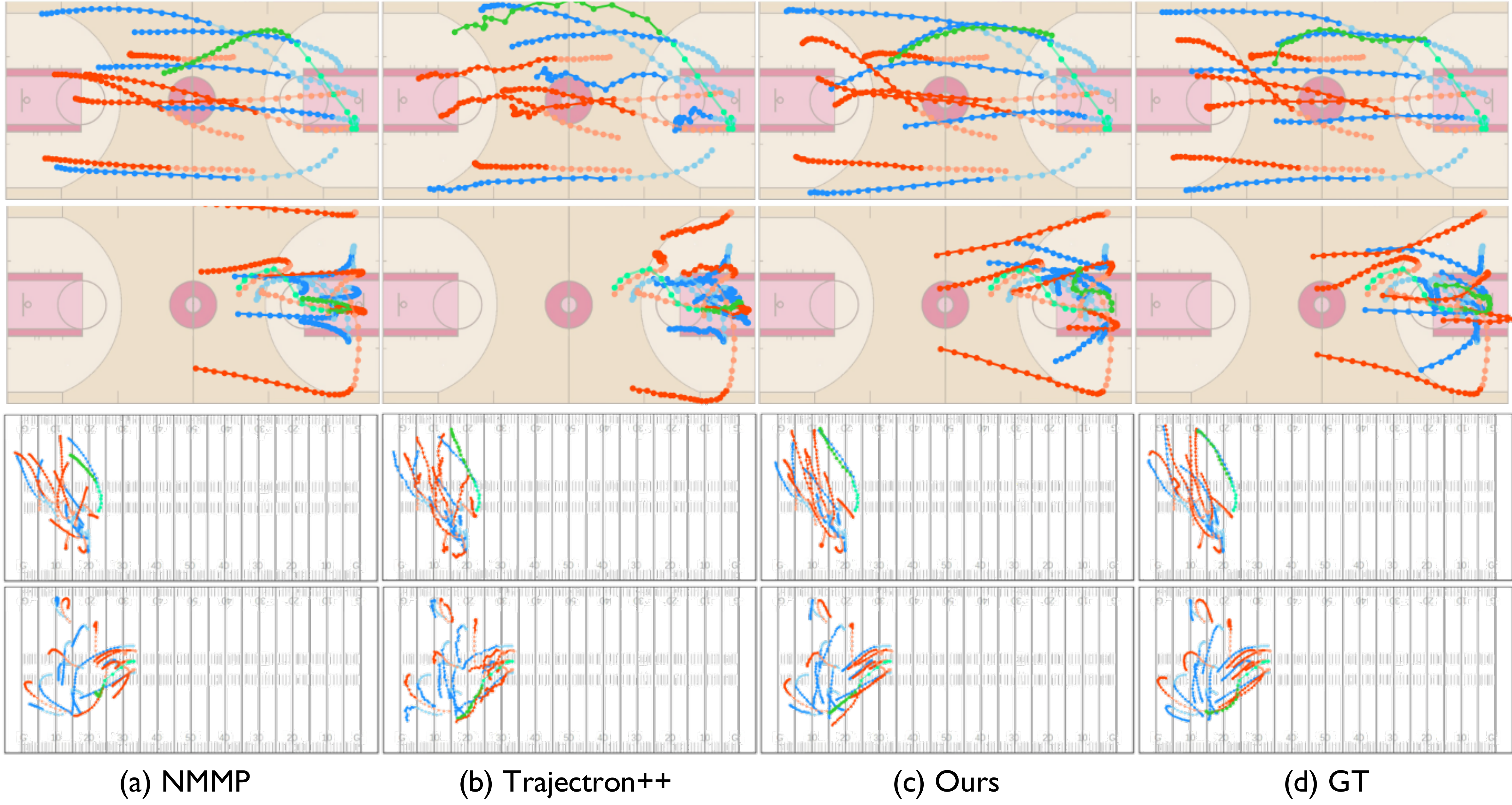}
\vspace{-2mm}
\caption{\small Visualization results on NBA and NFL Football datasets. The red/blue color represents players of two teams and the green color represents the basketball. Light color represents the past trajectory.}
\label{fig:nba}
\vspace{-2mm}
\end{figure*}

\subsubsection{Ablation on evolving gap \& decoding length}
TABLE \ref{table:timestamps} presents the effect of different evolving gap $\tau$ \& decoding length $T_{\rm D}$ in the~\texttt{DynGroupNet} on the NBA dataset. A lower encoding gap/decoding length indicates a more frequent change of the dynamic multiscale hypergraphs and $\tau=T_{\rm D}=20$ means using a static multiscale hypergraph. We see that i) using a dynamic multiscale hypergraph performs better than a static multiscale hypergraph since a dynamic multiscale hypergraph is capable to handle the time-varying interaction among agents; ii) too frequent evolving in a dynamic multiscale hypergraph led by a small evolving gap is unnecessary since interactions among agents rarely change very frequently across time; and iii) a moderate encoding gap/decoding length leads to the superior performance. 
 
\subsubsection{Ablation on dynamic embedding evolvement}
Fig. \ref{fig:ablation_dfe} presents the effect of dynamic embedding evolvement in the~\texttt{DynGroupNet} on both the NBA and the NFL Football datasets. Shadow/solid bars represent without/with dynamic embedding evolvement. We see that with the proposed dynamic embedding evolvement, the performance has a significant improvement because previous agents and their interaction embeddings will impact their current movement, which is modeled by the dynamic embedding evolvement.

\subsubsection{Ablation on~\texttt{DynGroupNet} modules}
We perform extensive ablation studies on the NBA dataset to investigate the contribution of the key technical component~\texttt{DynGroupNet}; see TABLE \ref{table:module}. We apply four other feature extraction modules. The `MLP' means that we use a multi-layer perceptron. The `ATT' and `NMMP' means that using the non-local attention mechanism in \cite{wang2018non} and using the graph neural message passing in \cite{hu2020collaborative} which models the pair-wise interactions between two agents. The 'EdgeAgg' means using an element-wise sum as the aggregation operation to combine neighboring features in \cite{salzmann2020trajectron++}. We see that the proposed~\texttt{DynGroupNet} leads to superior performance comparing to the multi-layer perceptron, the attention mechanism, the graph-based mechanism and the edge aggregation mechanism.

\begin{table}[!t]
    \caption{\small  Ablation studies of different feature extraction modules on the NBA dataset. We report minADE$_{20}$/minFDE$_{20}$ (meters). Our method achieves the best performance.}
    \vspace{-2mm}
    \centering
    \setlength{\tabcolsep}{1.5mm}{\small
\begin{tabular}{c|cccc}
\hline
\hline
% \multicolumn{1}{c|}{$\mathcal{M}$} &
\multicolumn{1}{c|}{} & \multicolumn{4}{c}{Prediction time}  \\ \hline
Module & 1.0s & 2.0s & 3.0s & 4.0s \\
\hline
MLP  &  0.23/0.34 & 0.48/0.77 & 0.76/1.14 & 1.02/1.42 \\
% MLP   & ATT   &  0.36/0.51  &  0.65/1.03    &  0.94/2.46    & 1.23/1.91    \\
% MLP  & GroupNet & 0.36/0.51&0.66/1.03&0.95/1.47&1.23/1.91    \\
SocialPool     & 0.22/0.32 & 0.46/0.71 & 0.71/1.04 & 0.96/1.31  \\
NMMP & 0.24/0.35 & 0.49/0.73 & 0.75/1.05 & 1.00/1.30\\
% GroupNet  & MLP          &     0.35/0.49    &   0.62/0.95   &   0.88/1.32   & 1.14/1.70   \\
% GroupNet  & ATT &     0.35/0.49    &   0.63/0.95   &   0.88/1.32   & 1.14/1.70   \\
EdgeAgg &  0.22/0.32& 0.46/0.69 & 0.72/1.02 & 0.97/1.27 \\
DynGroupNet   &   \textbf{0.19}/\textbf{0.28}&\textbf{0.40}/\textbf{0.61}&\textbf{0.65}/\textbf{0.90}&\textbf{0.89}/\textbf{1.13} \\
         \hline
          \hline
\end{tabular}}
\label{table:module}
\end{table}

\setlength{\tabcolsep}{3pt}
\begin{table}[t]
\footnotesize
\centering
\renewcommand{\arraystretch}{1.2}
\caption{\small Ablation study of GMM $\&$ Multi-sampling (MS) $\&$ Prediction refinement (PR) in prediction system on the NBA and the NFL Football dataset. We report minADE$_{20}$/minFDE$_{20}$ (meters). All designs are beneficial.}
\vspace{-2mm}
\setlength{\tabcolsep}{1mm}{\small
\begin{tabular}{ccc|cc|cc}
\hline
\hline
\rule{0pt}{10pt} % change row height

\multirow{2}{*}{\makecell[c]{GMM}} &\multirow{2}{*}{\makecell[c]{MS}} & \multirow{2}{*}{\makecell[c]{PR}} & \multicolumn{2}{c|}{\makecell[c]{NBA}} & \multicolumn{2}{c}{NFL Football} \\
 &  & &2.0s & 4.0s & 2.0s & 3.2s \\
\hline
  &  && 0.62/0.95 & 1.13/1.69 &0.73/1.39&1.21/2.15\\
 \checkmark& & &0.43/0.67&0.93/1.27&0.69/1.13&1.12/1.71  \\
 \checkmark &  \checkmark &&0.42/0.63&0.90/1.16& 0.64/1.10&1.06/1.66 \\
  \checkmark &  \checkmark &\checkmark&\textbf{0.40}/\textbf{0.61}&\textbf{0.89}/\textbf{1.13}&\textbf{0.56}/\textbf{0.94}& \textbf{0.95}/\textbf{1.49} \\
\hline
\hline
\end{tabular}}
\label{table:ablation_sys}
\end{table}

\begin{figure}[t] 
\centering
\includegraphics[width=0.47\textwidth]{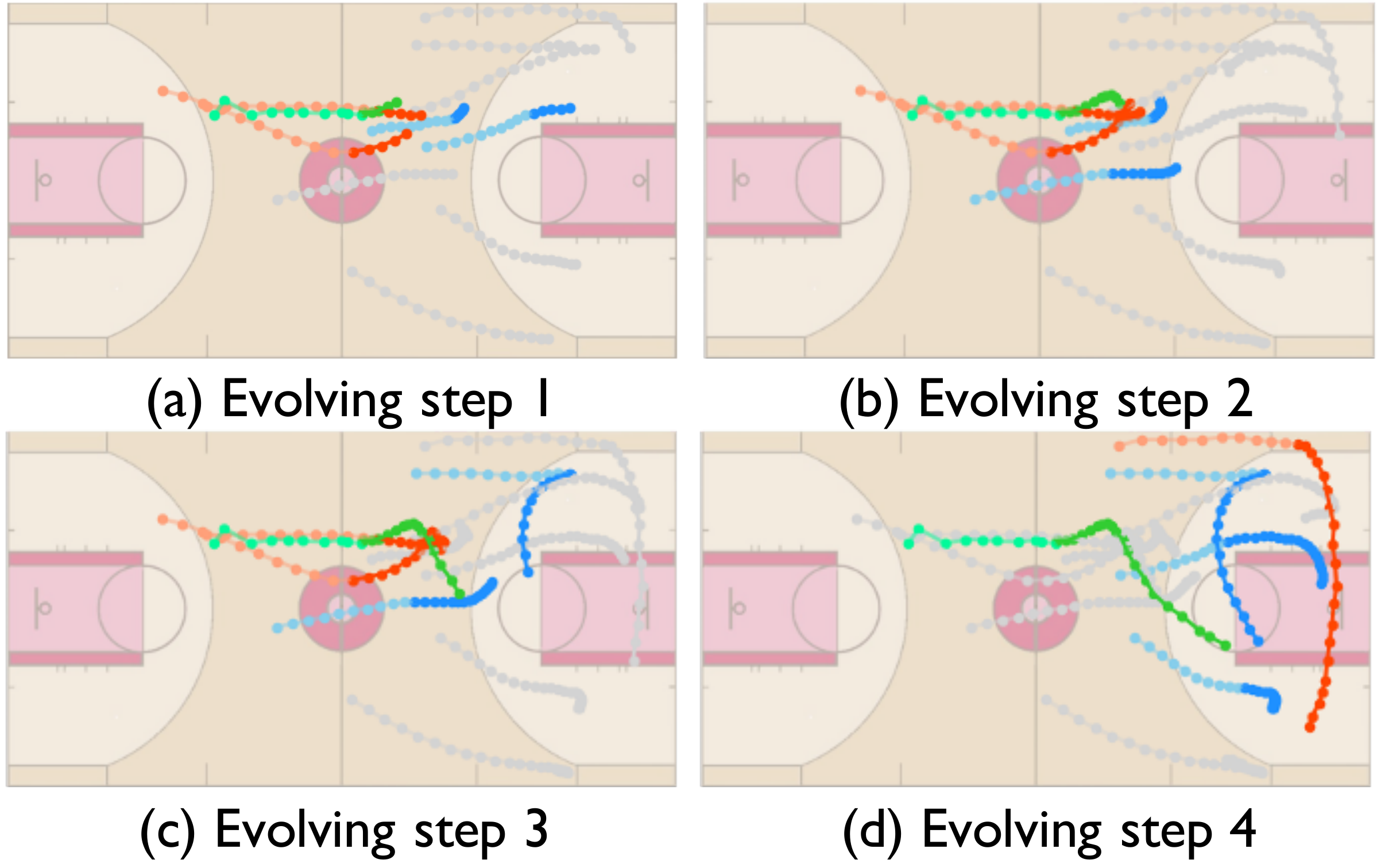}
\vspace{-2mm}
\caption{\small Visualization of time-varying groups. We visualize a group containing 5 agents including the ball of different evolving steps. Colored trajectories represent belonging to the group and gray trajectories represent not belonging to the group.}
\label{fig:nbagroup}
\end{figure}

\subsubsection{Ablation on prediction system designs}
We perform extensive ablation studies on both the NBA dataset and NFL Football dataset to investigate the contribution of technical designs in the prediction system including the bivariate gaussian mixture model~(GMM), multiple sampling~(MS) and prediction refinement~(PR); see TABLE \ref{table:ablation_sys}. We see that our proposed technical designs in the prediction system all contribute to promoting accurate prediction, reflecting the effectiveness of our prediction system.

\subsection{Visualizations}
\mypar{Visualization on prediction results}
Fig. \ref{fig:nba} compares the predicted trajectories of two baselines NMMP, Trajectron++, our~\texttt{DynGroupNet}(Ours) and ground-truth (GT) trajectories on both NBA dataset (row 1-2) and NFL Football dataset (row 3-4). We visualize the best-of-20 predicted trajectories evaluated by FDE. We see that i) our method produces more precise predictions than previous state-of-the art method NMMP, Trajectron++ on both two datasets; and ii) for the scenes of fierce confrontation between two teams (the second and fourth row), our method has a larger improvement. This is because our method captures more comprehensive interactions among players across time in complicated scenes.

\mypar{Visualization on group behaviors}
Fig. \ref{fig:nbagroup} presents the visualization of group behavior varying across time. Here we visualize a group containing 5 agents including the ball at different evolving steps. Agents in the group are plotted in colors. We see that i) at first, the ball is carried by a red player and the group contains two red players who want to attack and two blue players who want to defend; and ii) when the ball is passed away, the group changes to contain the red player who wants to receive the ball and three blue players who want to stop this ball-passing. This reflects that our method is capable to capture reasonable dynamic group behaviors.

% \subsubsection{Ablation on predefine interaction category numbers}
% (draw figure)

% \begin{table}[!t]
%     \caption{\small Ablation studies of different interaction category numbers on the NBA dataset. We report minADE$_{20}$/minFDE$_{20}$ (meters). }
%     \centering
%     \setlength{\tabcolsep}{1.mm}{\small
% \begin{tabular}{c|cccc}
% \hline
% \hline
% % \multicolumn{1}{c|}{$\mathcal{M}$} &
% \multicolumn{1}{c|}{\makecell[c]{Interaction \\category numbers}} & \multicolumn{4}{c}{Prediction time}  \\ \hline
%  & 1.0s & 2.0s & 3.0s & 4.0s \\
% \hline
% 2  &   \\
% % MLP   & ATT   &  0.36/0.51  &  0.65/1.03    &  0.94/2.46    & 1.23/1.91    \\
% % MLP  & GroupNet & 0.36/0.51&0.66/1.03&0.95/1.47&1.23/1.91    \\
% 4 & \\
% 6    &  \\
% % GroupNet  & MLP          &     0.35/0.49    &   0.62/0.95   &   0.88/1.32   & 1.14/1.70   \\
% % GroupNet  & ATT &     0.35/0.49    &   0.63/0.95   &   0.88/1.32   & 1.14/1.70   \\
% 8   &    \\
% 10   &   \textbf{0.19}/\textbf{0.28}&\textbf{0.41}/\textbf{0.62}&\textbf{0.65}/\textbf{0.91}&\textbf{0.89}/\textbf{1.13} \\
% 20   &    \\
%          \hline
%           \hline
% \end{tabular}}
% \label{table:sys}
% \end{table}

\section{Conclusion}
\label{sec:conclusion}
In this paper, we propose \texttt{DynGroupNet}, which promotes a more comprehensive interaction modeling from three aspects: i) designing a multiscale hypergraph to capture both pair-wise and group-wise interactions; ii) evolving the multiscale hypergraph to a dynamic multiscale hypergraph to capture dynamic interactions; and iii) designing a three-element representation format of interaction embedding, including neural interaction strength, category and per-category function. We introduce a prediction system cooperated with \texttt{DynGroupNet} as the core to predict dynamically-feasible future trajectories. In the prediction system, we design a GMM model, a multiple sampling training strategy, and a prediction refinement. We conduct extensive experiments on synthetic physics simulations and real-world datasets. Experiments reveal the ability of relational reasoning and the effectiveness of \texttt{DynGroupNet} and the prediction system.

% or


\appendices

% \section{}

\ifCLASSOPTIONcompsoc
  \section*{Acknowledgments}
\else
  \section*{Acknowledgment}
\fi

This research is partially supported by the National Key R\&D Program of China under Grant 2021ZD0112801, National Natural Science Foundation of China under Grant 62171276, the Science and Technology Commission of Shanghai Municipal under Grant 21511100900 and CCF-DiDi GAIA Research Collaboration Plan 202112.

\ifCLASSOPTIONcaptionsoff
  \newpage
\fi



%\bibliographystyle{IEEEtran}

% {\small
% \bibliographystyle{plainnat}
% \bibliography{egbib}
% }
\bibliographystyle{IEEEtran}
\bibliography{egbib}

% \end{thebibliography}

\begin{IEEEbiography}[{\includegraphics[width=1in,height=1.25in,clip,keepaspectratio]{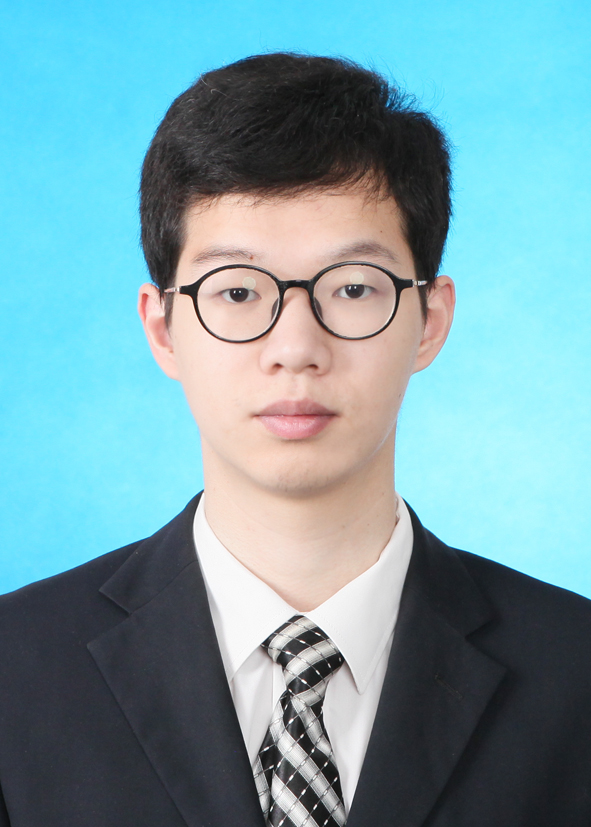}}]{Chenxin Xu} received the B.E. degree in information engineering from Shanghai Jiao Tong University, Shanghai, China, in 2019. He is working toward the joint
Ph.D. degree at Cooperative Medianet Innovation Center in Shanghai Jiao Tong University and at Electrical and Computer Engineering in National University of Singapore since 2019. His research interests include computer vision, machine learning, graph neural network, and multi-agent prediction. 
\end{IEEEbiography}
\vspace{-1mm}
\begin{IEEEbiography}[{\includegraphics[width=1in,height=1.25in,clip,keepaspectratio]{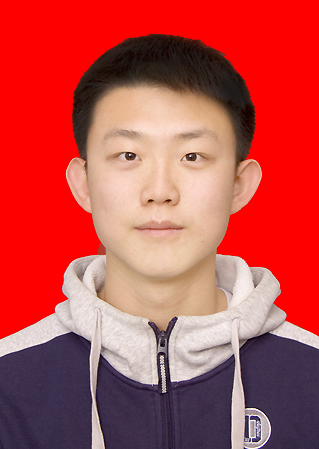}}]{Yuxi Wei} is an intern at Cooperative Medianet Innovation Center in Shanghai Jiao Tong University. He is working toward B.E. degree in artificial intelligence in Shanghai Jiao Tong University. His research interests include machine learning, graph neural network, and multi-agent prediction. 
\end{IEEEbiography}
\vspace{-1mm}
\begin{IEEEbiography}[{\includegraphics[width=1in,height=1.25in,clip,keepaspectratio]{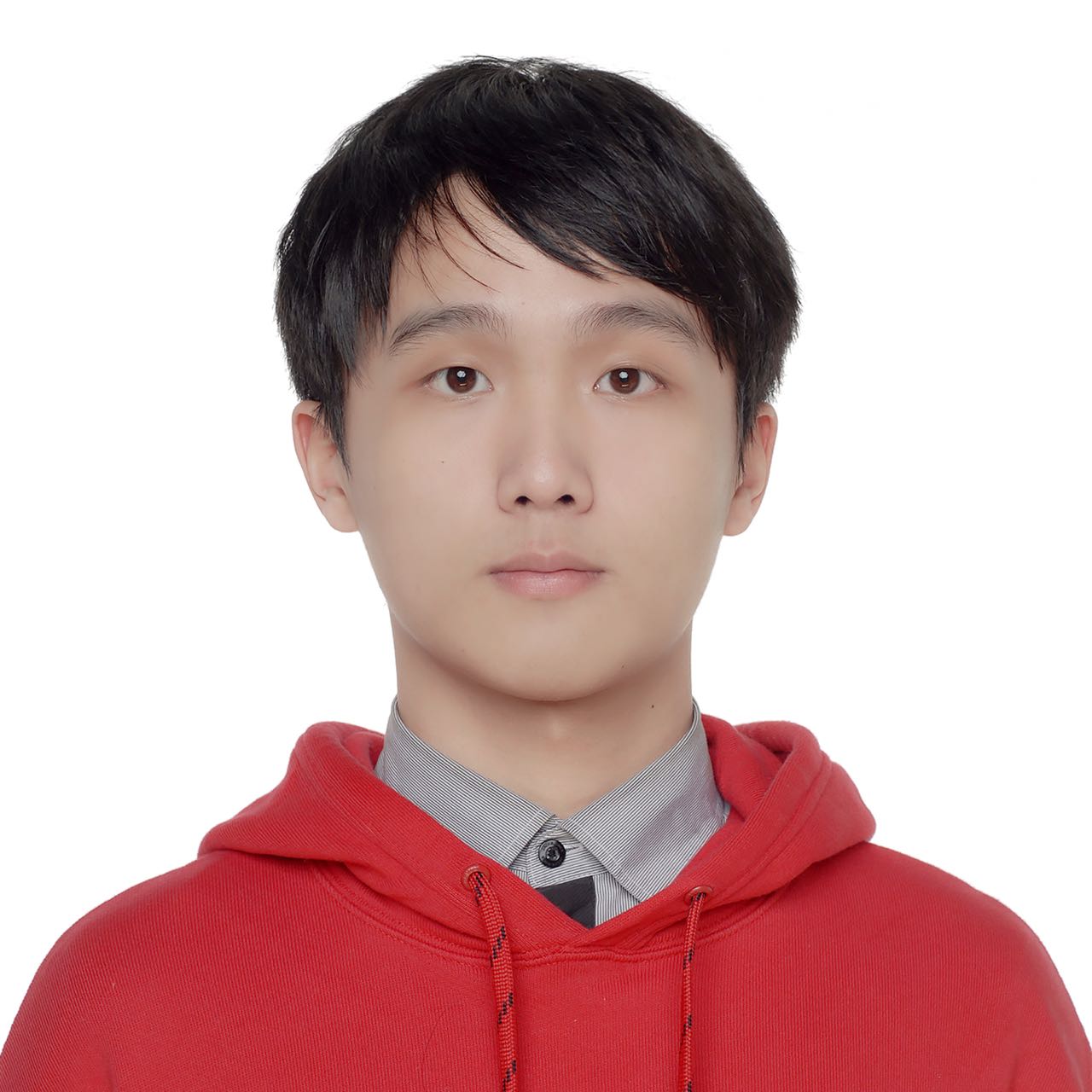}}]{Bohan Tang} received the B.E. degree in information engineering from Shanghai Jiao Tong University, Shanghai, China, in 2021. He is working toward the Ph.D. degree at the Oxford-Man Institute and the Department of Engineering Science in University of Oxford since 2021. His research interests include machine learning, graph neural network, graph learning, hypergraph learning and uncertainty estimation.
\end{IEEEbiography}
\vspace{-1mm}

\begin{IEEEbiography}[{\includegraphics[width=1in,height=1.25in,clip,keepaspectratio]{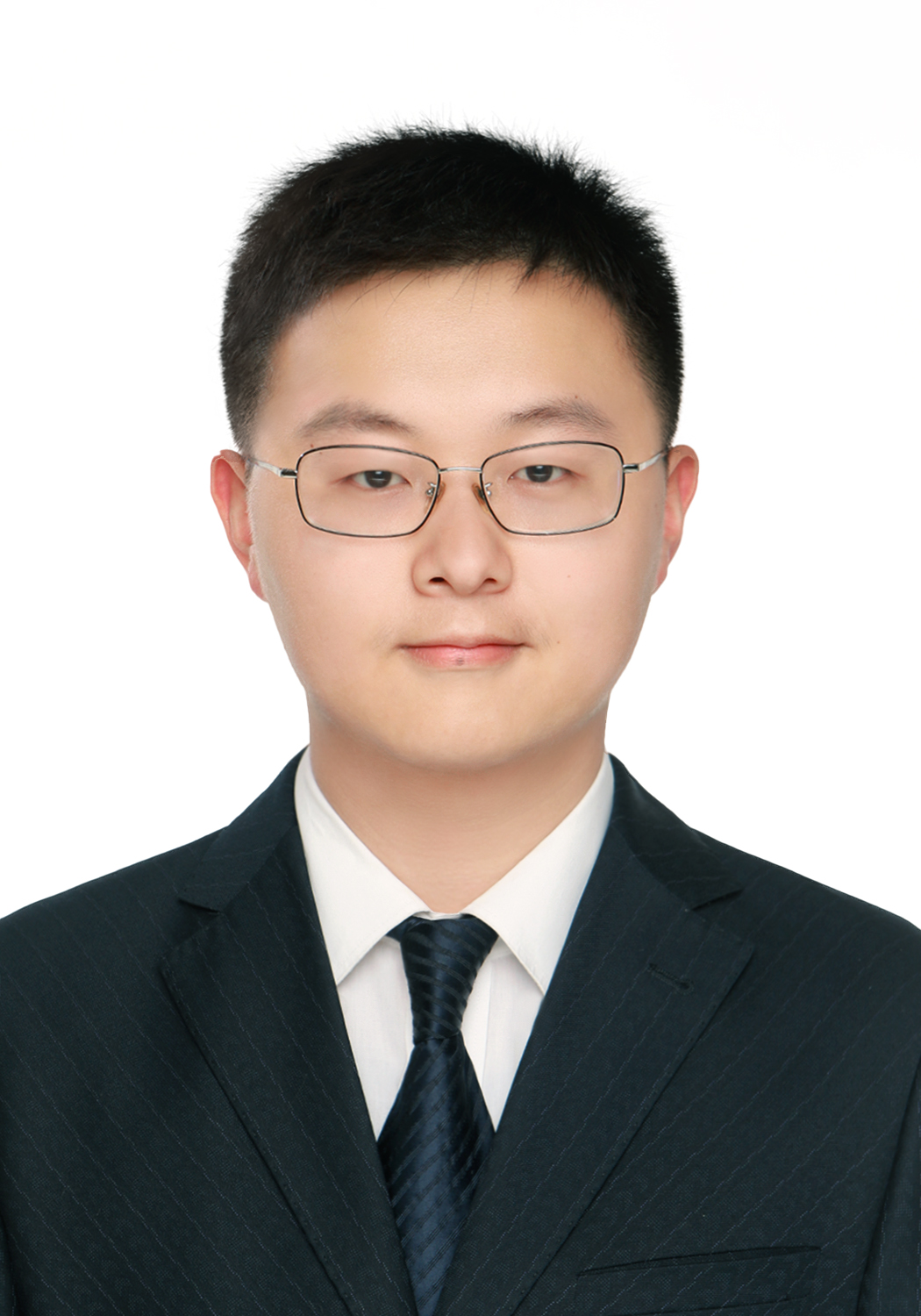}}]{Sheng Yin} is an intern at Cooperative Medianet Innovation Center in Shanghai Jiao Tong University. He is working toward B.E. degree in IEEE Pilot Class in Shanghai Jiao Tong University. His research interests include machine learning, federated learning, and graph learning.
\end{IEEEbiography}
\vspace{-1mm}

%\newpage

\begin{IEEEbiography}[{\includegraphics[width=1in,height=1.25in,clip,keepaspectratio]{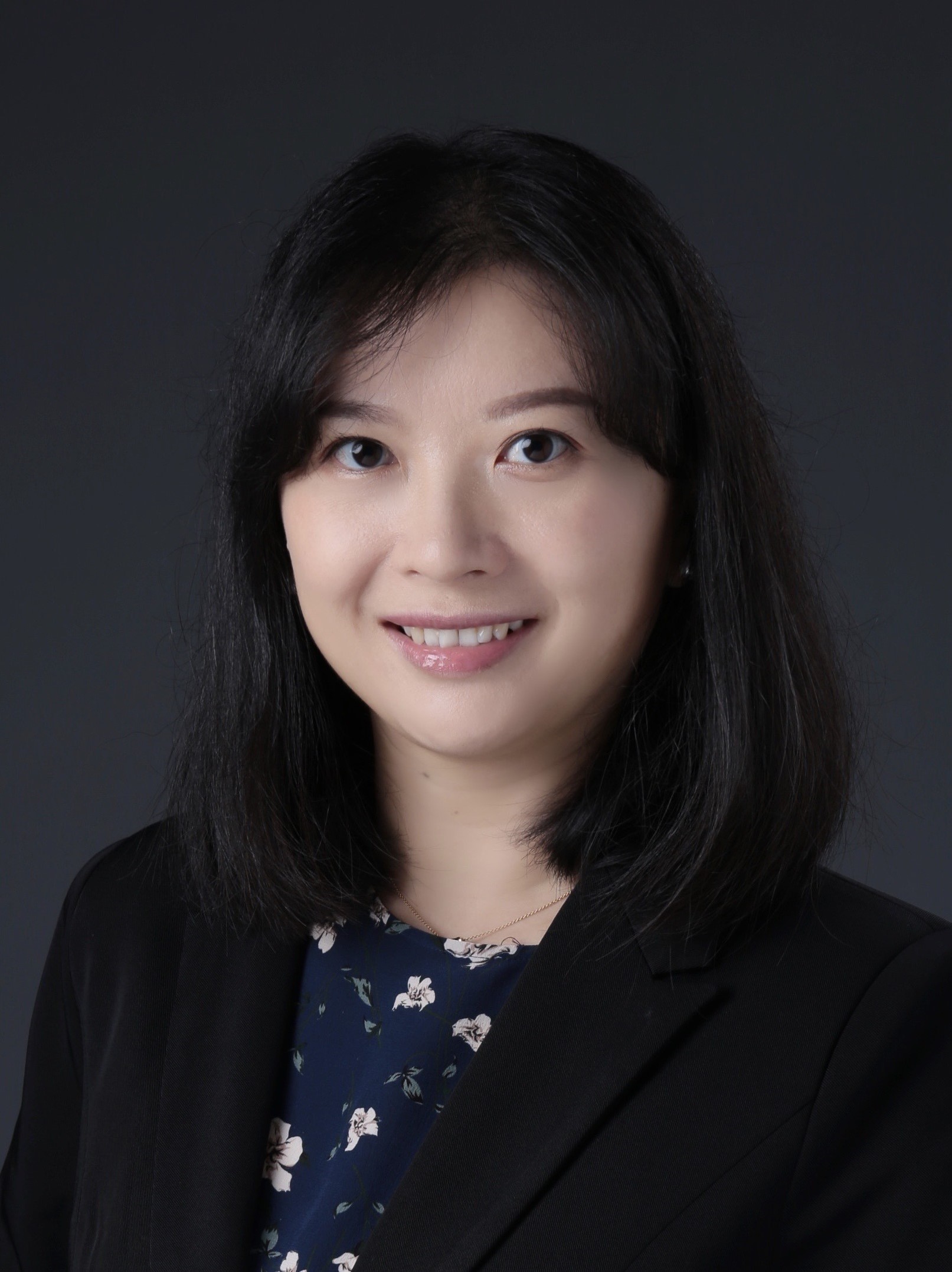}}]{Ya Zhang} is currently a professor at the Cooperative Medianet Innovation Center, Shanghai Jiao Tong University. Her research interest is mainly in machine learning with applications to multimedia
and healthcare. Dr. Zhang holds a Ph.D. degree in Information Sciences and Technology from Pennsylvania State University and a bachelor’s degree from Tsinghua University in China. Before joining Shanghai Jiao Tong University, Dr. Zhang was a research manager at Yahoo! Labs, where she led an R\&D team of researchers with strong backgrounds in data mining and machine learning to improve the web search quality of Yahoo international markets. Prior to joining Yahoo, Dr. Zhang was an assistant professor at the University of Kansas with a research focus on
machine learning applications in bioinformatics and information retrieval. Dr. Zhang has published more than 70 refereed papers in prestigious international conferences and journals, including TPAMI, TIP, TNNLS, ICDM, CVPR, ICCV, ECCV, and ECML. She currently holds 5 US patents and 4 Chinese patents and has 9 pending patents in the areas of multimedia analysis. She was appointed the Chief Expert for the ’Research of Key Technologies and Demonstration for Digital Media Self-organizing’ project under the 863 program by the Ministry of Science and Technology of China. She is a member of IEEE.
\end{IEEEbiography}
\vspace{-1mm}

\begin{IEEEbiography}[{\includegraphics[width=1in,height=1.25in,clip,keepaspectratio]{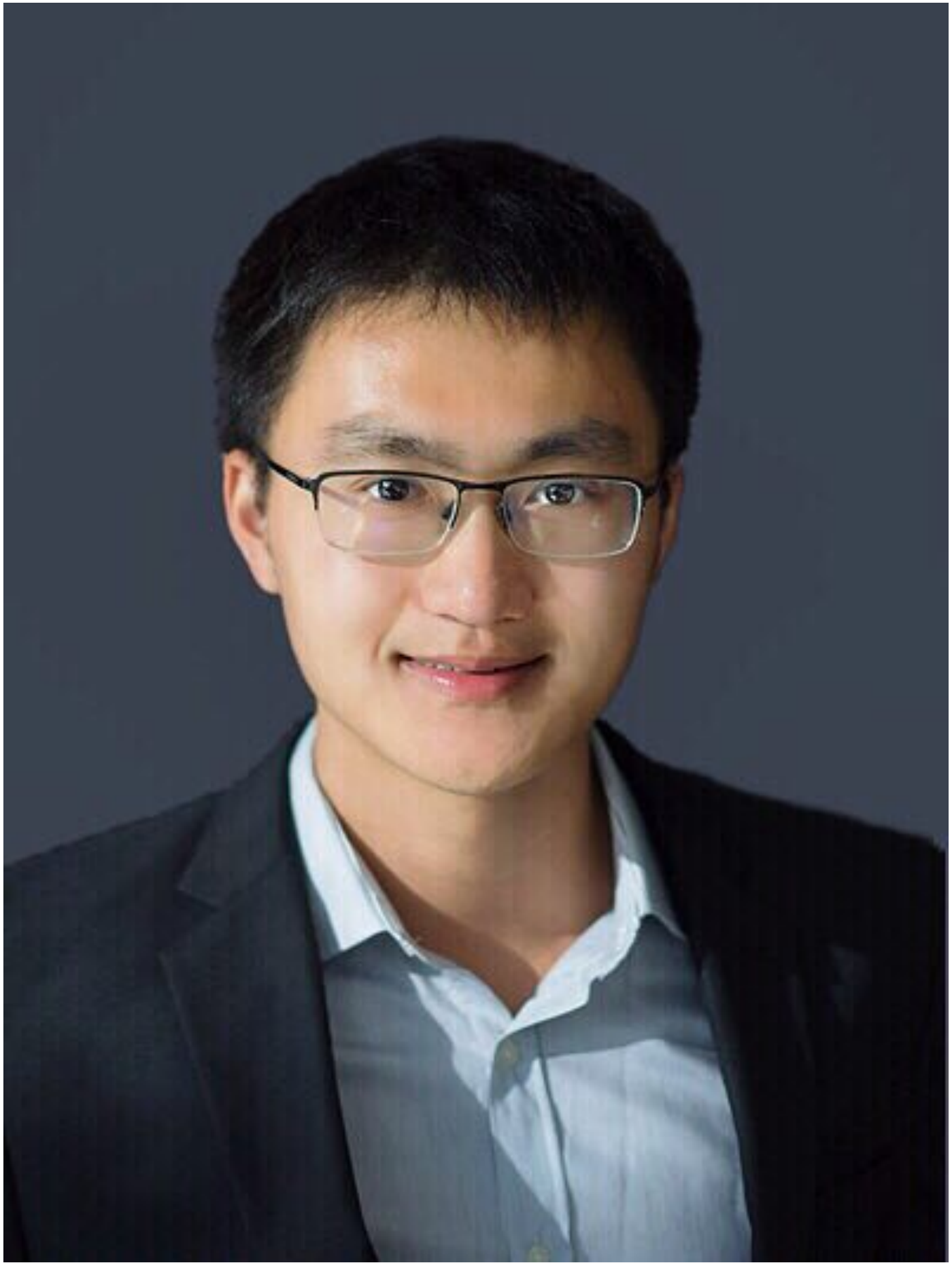}}]{Siheng Chen} is a tenure-track associate professor of Shanghai Jiao Tong University. Before joining Shanghai Jiao Tong University, he was a research scientist at Mitsubishi Electric Research Laboratories (MERL), and an autonomy engineer at Uber Advanced Technologies Group (ATG), working on the perception and prediction systems of self-driving cars. Before joining industry, Dr. Chen was a postdoctoral research associate at Carnegie Mellon University. Dr. Chen received his doctorate in Electrical and Computer Engineering from Carnegie Mellon University in 2016, where he also received two master degrees in Electrical and Computer Engineering (College of Engineering) and Machine Learning (School of Computer Science), respectively. He received his bachelor’s degree in Electronics Engineering in 2011 from Beijing Institute of Technology, China. Dr. Chen's work on sampling theory of graph data received the 2018 IEEE Signal Processing Society Young Author Best Paper Award. His co-authored paper on structural health monitoring received ASME SHM/NDE 2020 Best Journal Paper Runner-Up Award and another paper on 3D point cloud processing received the Best Student Paper Award at 2018 IEEE Global Conference on Signal and Information Processing. Dr. Chen contributed to the project of scene-aware interaction, winning MERL President's Award. His research interests include graph neural networks, autonomous driving and collective intelligence.
\end{IEEEbiography}
\vspace{-1mm}

\appendices
\section{Generation Details of Simulation Datasets}
\subsection{Ability to capture dynamic group behaviors}
In this simulation, we have 6 particles with a same mass in an x-y plane. Initially, 6 particles are divided into three groups. Each group contains two particles connected by a light bar. The length of the light bar is in the range [0.3,0.6]. The light bar will rotate with a random angular velocity ranging in $[\pi/15,\pi/4]$ and translate with a random translational velocity ranging in $[0.05,0.2]$. At one moment, two particles of different light bars will collide. After collision, the two particles merge as one and thus two light bars will also merge as a "L" shape. The merged "L" shape bar will have a new angular velocity and a new translational velocity following the conservation of angular momentum principle. We predict the particle states at the future 15 timestamps based on the observation of 10 timestamps and generated 5k, 2k samples for training and testing.

\subsection{Ability to reason dynamic interaction category}
In this simulation, we have 3 particles with the same mass in an x-y plane. Initially, 3 particles are connected by a "Y-shape" light bar. The light bar will rotate with a random angular velocity ranging in $[\pi/10,\pi/5]$ and translate with a random translational velocity ranging in $[-0.1,-0.2]$ in the x-direction and $[0,0.2]$ in the y-direction. The length of the light bar (from center to particle) is ranging in $[0.5,1]$. The initial center of the light bar is ranging in $[2,4]$ on the x-axis and $[0,3]$ on the y-axis. We predict the particle states at the future 20 timestamps based on the observation of 10 timestamps and generated 5k and 2k samples for training and testing.

\subsection{Ability to reason dynamic interaction strength}
In this simulation, we have two charged particles with the same mass in an x-y plane. The charged particles interact under Coulomb force:$F=C\cdot sign(q_1\cdot q_2)\frac{(r_1-r_2)}{||r_1-r_2||^3}$, where C is a constant and $q_1,q_2$ is the charged quantity $r_1,r_2$ is the position. One particle is fixed at the origin point, carrying a random initialized positively charged quantity ranging in $[3,16]$. Another particle starts from the position $(0,3)$ with a fixed negative charged quantity of $1$. The moving particle has an initial velocity of $(-1.5,-1)$ in $(x,y)$  direction. The constant C is set to be $0.4$. When the particle moves, it will encounter an air resistance which is proportional to its speed and the scale factor is $0.2$. We predict the particle states at the future 25 timestamps based on the observation of 15 timestamps and generated 5k and 2k samples for training and testing.

\end{document}